\documentclass{article}

\usepackage{microtype}
\usepackage{graphicx}
\usepackage{subcaption}
\usepackage{booktabs} % for professional tables

\usepackage{hyperref}

\usepackage[accepted]{icml2026}

\usepackage{amsmath,amsfonts,bm}

\def\eqref#1{equation~\ref{#1}}
\def\1{\bm{1}}

\DeclareMathAlphabet{\mathsfit}{\encodingdefault}{\sfdefault}{m}{sl}
\SetMathAlphabet{\mathsfit}{bold}{\encodingdefault}{\sfdefault}{bx}{n}

\newcommand{\E}{\mathbb{E}}

\usepackage{amsmath}
\usepackage{amssymb}
\usepackage{mathtools}
\usepackage{amsthm}
\usepackage{hyperref}
\usepackage{url}
\usepackage{booktabs}       % professional-quality tables
\usepackage{amsfonts}       % blackboard math symbols
\usepackage{nicefrac}       % compact symbols for 1/2, etc.
\usepackage{microtype}      % microtypography
\usepackage{xcolor}         % colors
\usepackage{graphicx}
\usepackage{multirow}
\usepackage{array}
\usepackage{wrapfig}
\usepackage{amsmath}
\usepackage{paralist} 
\usepackage{float}
\usepackage{tabularx}
\usepackage{booktabs}
\usepackage{graphicx}
\usepackage{enumitem}
\usepackage{caption}
\usepackage{longtable}
\usepackage{subcaption}
\usepackage{amsmath,amssymb}
\usepackage{adjustbox}
\newcommand{\calO}{\mathcal{O}}
\newcommand{\calP}{\mathcal{P}}

\usepackage[capitalize,noabbrev]{cleveref}

\theoremstyle{plain}

\theoremstyle{definition}

\theoremstyle{remark}

\usepackage{xcolor}

\newcommand{\updatetext}[1]{\textcolor{black}{#1}}
\newcommand{\updatetwotext}[1]{\textcolor{black}{#1}}

\usepackage[textsize=tiny]{todonotes}

\icmltitlerunning{Moving Out: Physically-grounded Human-AI Collaboration}

\begin{document}

\twocolumn[
  \icmltitle{Moving Out: Physically-grounded Human-AI Collaboration}

\begin{icmlauthorlist}
  \icmlauthor{Xuhui Kang}{uva}
  \icmlauthor{Sung-Wook Lee}{uva}
  \icmlauthor{Haolin Liu}{uva}
  \icmlauthor{Yuyan Wang}{uva}
  \icmlauthor{Yen-Ling Kuo}{uva}
\end{icmlauthorlist}

\icmlaffiliation{uva}{University of Virginia, Charlottesville, Virginia, USA}

\icmlcorrespondingauthor{Xuhui Kang}{xuhui@virginia.edu}
\icmlcorrespondingauthor{Yen-Ling Kuo}{ylkuo@virginia.edu}

\icmlkeywords{Physical AI, Embodied Agents, Human-AI Collaboration, Multiagent Reinforcement Learning, Imitation Learning}

  \vskip 0.3in
]



\printAffiliationsAndNotice{}  % no special notice (required even if empty)

\begin{abstract}
The ability to adapt to physical actions and constraints in an environment is crucial for embodied agents (e.g., robots) to effectively collaborate with humans.
Such physically grounded human-AI collaboration must account for the increased complexity of the continuous state-action space and constrained dynamics caused by physical constraints.
However, most existing collaboration benchmarks are discrete or do not consider physical attributes and constraints.
To address this, we introduce \textit{Moving Out}, a human-AI collaboration benchmark that resembles a wide range of collaboration modes affected by physical attributes and constraints, such as moving heavy items together and coordinating actions to move an item around a corner.
Moving Out consists of two \updatetext{challenges} and human-human interaction data to comprehensively evaluate models' abilities to adapt to diverse human behaviors and unseen physical attributes.
To give embodied agents the capability to collaborate with humans under physical attributes and constraints, we propose a novel method, BASS (Behavior Augmentation, Simulation, and Selection), to enhance the diversity of agents and their understanding of the outcome of actions.
We systematically compare BASS and state-of-the-art models in AI-AI and human-AI experiments, showing that BASS can effectively collaborate with both unseen AI and humans.
The project page is available at \href{https://live-robotics-uva.github.io/movingout_ai/}{https://live-robotics-uva.github.io/movingout\_ai/}.
\end{abstract}

\section{Introduction}
The ability to understand physical attributes (e.g., shapes, weights, etc.) or constraints (e.g., narrow paths, etc) and their impact on an agent's actions is core to collaboration in the physical world, such as moving a sofa together.
Humans can quickly recognize that one agent cannot move a heavy item alone to offer help, or coordinate the motion to move a table around a corner instead of canceling out the partner's effort.
This ability is critical for \emph{physically-grounded human-AI collaboration} where embodied agents (e.g., robots) need to collaborate with humans in the physical world.

\begin{figure}[!htp]
    \centering
    \vspace{-0.5em}
    \includegraphics[width=1.0\linewidth]{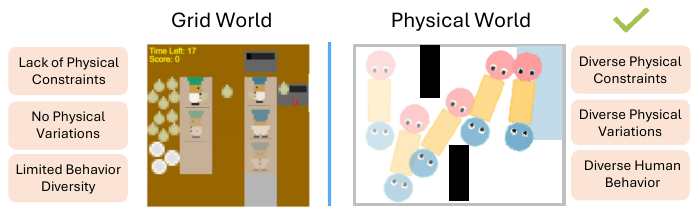}
    \vspace{-1.5em}
    \caption{Multiagent collaboration in a grid world (Overcooked-AI~\citep{carroll2019utility}) vs. in a physical world (this work). 
    %Physically grounded settings introduce diverse physical constraints, attributes, and continuous low-level actions, which are essential for developing collaborative AI that can operate in physical scenarios.
    }
    \vspace{-0.5em}
    \label{fig:motivation}
\end{figure}

Prior work~\citep{carroll2019utility, ng2022takes, DBLP:conf/nips/PapoudakisC0A21, puig2023habitat, christianos2020shared, duconstrained} has explored human-AI collaboration but the environment and agent actions are discrete, symbolic, or task-level, which often have simplified interaction dynamics compared to the interactions in a physical environment.
In physical environments, agents have a continuous state–action space where object interactions, motions, and task outcomes are affected by physical attributes and constraints such as mass, shape, and contact dynamics.
This usually results in an increased complexity of interaction dynamics.
As shown in Fig.~\ref{fig:motivation}, in a grid world, there are only a few fixed positions to pass an item to another agent and passing can complete in one step; in a physical world, while the object shape and the narrow passage restrict the movement, there are still a large number of rotations or ways of holding objects the agent needs to reason about in order to achieve success collaborations.
Here, we create \textit{Moving Out}, a novel benchmark inspired by the Moving Out game~\citep{MovingOut}, to study how embodied agents can collaborate with humans under these increased interaction dynamics.
Moving Out is built on top of a 2D physics engine with realistic rigid-body physics simulations.
Agents are fully observable and self-propelled to maneuver in the environment, and can move the objects with varying degrees of difficulty depending on object sizes, shapes, and mass.
Agents need to work together to move all objects to the goal regions.
The environment layouts bring additional physical constraints.
Based on the layouts, the agents will need to employ a mix of collaboration modes to succeed.

While AI-AI collaboration can achieve strong collaborative performance through methods like self-play~\citep{tesauro1994td}, the resulting AI agents often struggle to adapt to human partners who exhibit diverse behaviors~\citep{carroll2019utility}.
This is particularly pronounced in the physical world, where minor variations in human actions, such as rotation angles or applied forces, can significantly affect outcomes and interaction dynamics.
To truly measure whether an agent can engage in physically-grounded collaboration, we propose two \updatetext{challenges}.
The first requires the agent to play against unseen human behavior.
We collected over 1,000 pairs of human demonstrations on maps with fixed physical attributes from 36 human participants.
These demonstrations capture a wide range of behaviors for an identical set of tasks.
We split the data by participants into two disjoint splits so agents trained on one split have never seen the behavior of another and must adapt when they encounter the unseen behavior.
The second requires the agent to generalize to unseen physical constraints.
We collected 720 pairs of demonstrations from 4 experts on maps with randomly sampled object attributes, such as mass, size, and shape.
The evaluation maps contain the attributes that are not in the dataset.
So, an agent must understand the meaning of physical constraints to interpret the other agent's behavior correctly to collaborate.
%Together, these tasks provide a framework for testing the adaptability and generalization of embodied agents in diverse, physically grounded settings.

%To further address the challenges of  diverse behavior in the continuous state-action space and constrained transitions in physical environments, 
To address these challenges, we propose BASS (Behavior Augmentation, Simulation, and
Selection), a novel human-AI collaboration method which significantly outperforms prior works.
BASS first augments the behavior to enhance the diversity of the agent's collaborative partners.
Unlike single agent scenarios, random changes of agent trajectories cannot work because they can lead to incompatible behavior with the partner.
Our behavior augmentation strategy swaps the partner's states only when an agent's start and end poses in one sub-trajectory match the sub-trajectory in another interaction.
This encourages the agent to generate consistent behavior when the partner's behavior has variations.
BASS then trains a dynamics model of agent interactions so the agent can simulate the outcome of an action while considering the possible partner actions.
BASS uses the predicted states to score action candidates, allowing the agent to select actions that are more effective for collaboration, given the physical constraints.
We evaluate BASS in the Moving Out challenges.
We show that BASS outperforms baselines across key metrics such as task completion and waiting time.
Our user study further evaluated BASS against human participants, demonstrating its effectiveness in coordinating and assisting humans.

In summary, we make the following contributions: (1) We introduce \textit{Moving Out}, a continuous environment for physically grounded human-AI collaboration. (2) We propose two \updatetext{challenges} and collect a human dataset to examine how human behavior and physical constraints impact collaboration. It is the first benchmark with human-collected datasets designed to study continuous, low-level motion control. (3) We develop \textit{Behavior Augmentation, Simulation, and Selection} (BASS), which significantly improves human-AI collaborative performance in physical worlds.  

% \vspace{-1em}
\section{Related Work}
% \vspace{-1em}
\textbf{Environments for Human-AI Collaboration}~
Several multi-agent environments~\citep{leibo2021meltingpot,terry2020arcade} have been proposed for multi-agent reinforcement learning (RL), but many are competitive rather than cooperative.
For human-AI collaboration, prior environments largely adopt symbolic or discrete action space~\citep{carroll2019utility,christianos2020shared,bard2020hanabi} or social settings~\citep{puig2020watch,cao2024smart}.
%, e.g., OvercookedAI~\citep{carroll2019utility}, LBF and RWARE~\citep{christianos2020shared}, Hanabi~\citep{bard2020hanabi}, or social settings like Watch and Help~\citep{puig2020watch} and Smart Help~\citep{cao2024smart}.
%, and ~\citep{zhang2024building}. 
While these settings are useful for studying coordination, they lack rich physical constraints and embodied teamwork. 
Other efforts, including It Takes Two~\citep{ng2022takes}, HumanTHOR~\citep{wang2024demonstratinghumanthorsimulationplatform}, and Habitat 3.0~\citep{puig2023habitat}, incorporate more realistic simulation.
However, It Takes Two provides only a single, highly simplified task, while HumanTHOR and Habitat 3.0 focus primarily on navigation or high-level task coordination.
In contrast, Moving Out provides continuous control, diverse physical attributes, and multiple collaboration modes, enabling the study of how AI can adapt to human behaviors under physical constraints. For a summarized comparison, see Appendix~\ref{appx:comparsion}.

\textbf{Learning Human-AI Collaboration Policy}~
Behavior Cloning (BC) is a common paradigm for learning policies from human demonstrations, typically using MLPs~\citep{rumelhart1985learning}, GRUs~\citep{cho2014learning}, or diffusion models~\citep{chi2023diffusion}. 
Beyond BC, several works extend it by predicting and scoring future states or trajectories~\citep{wang2024driving,Kang_2025_WACV,Yuan_2021_ICCV,zhao2021tnt,9143393}.
%, such as future-state prediction~, interactive agent forecasting~\citep{Yuan_2021_ICCV}, and trajectory-level scoring~\citep{zhao2021tnt,9143393}. 
RL further enhances collaboration via self-play~\citep{tesauro1994td} and population-based training~\citep{jaderberg2017population}, encouraging diverse behaviors for zero-shot coordination~\citep{carroll2019utility,strouse2021collaborating,yan2023efficient,li2023cooperative,yu2023learning,zhao2023maximum,sarkar2023diverse}. 
Standard multi-agent RL methods~\citep{yu2022surprising,lowe2017multi}, have also been applied. 
However, most RL methods rely solely on self-play without human data; recent work integrates BC-trained models into RL to align agents with human behavior~\citep{liang2024learning,carroll2019utility}.

\textbf{Evaluating Human-AI Collaboration}~
Research on human-AI collaboration has focused on evaluating AI agents in various settings.
\citet{attig2024more} define evaluation criteria beyond task performance, incorporating aspects like trust and perceived cooperativity .
In AI-assisted decision-making, \citet{vollmuth2023artificial} directly computes the accuracy of AI decisions.
Some works~\citep{tylkin2021learning,strouse2021collaborating,sarkar2023diverse} focus on training RL agents to adapt to diverse partners and evaluate the agents by the score when playing with humans.
Several works~\citep{sarkar2023diverse, attig2024more, siu2021evaluation,mckee2024warmth,hoffman2019evaluating} design questionnaires to evaluate aspects like human-like, trustworthiness, and fluency.

% \vspace{-1em}
\section{Problem Definition}

%\vspace{-1em}
%Follow by ~\citep{sarkar2023diverseconventionshumanaicollaboration},
We model human-AI collaboration as a decentralized Markov Decision Process (Dec-MDP)~\citep{beynier2013dec,boutilier1996planning}, defined as %a tuple 
 \(\mathcal{M} = (\mathcal{S}, \mathcal{A}, \calP,  r, \calO, \gamma, T)\), where \(\mathcal{S}\) is the joint state space, and \(\mathcal{A} = \mathcal{A}^i \times \mathcal{A}^j\) is the joint action space of the two agents.
The transition function $\calP: \mathcal{S} \times \mathcal{A} \times \mathcal{S} \rightarrow [0,1]$ is the probability of getting the next state given a current state and a joint action.
The reward function \(r: \mathcal{S} \times \mathcal{A} \to \mathbb{R}\) specifies the reward received for each state-joint-action pair.
The observation function $\calO: \mathcal{S} \rightarrow \calO^i \times \calO^j$ generates an observation for each agent for a given state.
The observation of each agent makes the state jointly fully observable.
The discount factor \(\gamma \in [0, 1]\) determines the importance of future rewards, and \(T\) is the time horizon of the task.

At each time \(t\), the environment is in a state \(s_t \in \mathcal{S}\).
Agent $i$ observes \(o_t^i \in \mathcal{O}\), where \(\mathcal{O}\) is the observation space derived from \(s_t\), and selects an action \(a_t^i \in \mathcal{A}^i\) according to its policy \(\pi^i: \mathcal{O} \to \mathcal{A}^i\).
The joint action \(a_t = (a_t^i, a_t^j)\) transitions the environment deterministically to a new state \(s_{t+1} \sim \calP(\cdot|s_t, a_t)\).
The trajectory of an episode is defined as \(\tau = (s_0, a_0, s_1, \dots, s_{T-1}, a_{T-1}, s_T)\), and the discounted return for the trajectory is:
\(
R(\tau) = \sum_{t=0}^{T-1} \gamma^t r(s_t, a_t).
\)
The objective of each agent is to maximize the expected return \(J(\pi^i, \pi^j) = \sum_{\tau} R(\tau)\)
where the return is evaluated over the trajectories induced by the policies \((\pi^i, \pi^j)\).

\textbf{Challenges when Collaborating with Humans}~
When one of the agents is a human, the human agent may have diverse behavior~\citep{carroll2019utility}.
The AI agent must adapt its policy \(\pi^i\) to a wide range of potential human policies \(\pi^j\).
At inference time, we assume that the real human policy \(\pi^j\) is drawn from an unknown human policy distribution $\mathcal{D}$.
Thus, the AI agent's optimal policy is:
\[
\pi^i_{\star} = \arg\max_{\pi^i} \E_{\pi^j \sim \mathcal{D}}\E_{\tau \sim (\pi^i, \pi^j)}\left[R(\tau)\right]
\]
% \vspace{-1em}
where $\E_{\tau \sim (\pi^i, \pi^j)}$ denotes the expectation over $\tau$ where the actions are drawn from $\pi^i$ and $\pi^j$ respectively.
Since the ground-truth distribution $\mathcal{D}$ is unknown, the AI must use limited data to generalize across diverse human strategies.

The physical embodiment of agents and the physical environment introduce significant challenges for this human-AI collaboration framework.
First, the continuous variables, e.g., positions and directions, increase the number of configurations in the state space.
For example, there are multiple configurations that an agent can take to rotate an object together.
The AI agent must optimize its policy under diverse human behaviors while ensuring robustness across a continuous and high-dimensional state space.
%Achieving this goal requires learning policies that adapt dynamically to both the variability of human actions and the complexities of physical environments, ensuring task success across a broad range of collaborative scenarios.
Second, the state space \(\mathcal{S}\) also includes continuous physical variables such as object positions, orientations, and attributes (e.g., shape, size, and mass), which can create constraints to limit the feasible state transitions \(\calP\).
%
%the state space \(\mathcal{S}\) incorporates continuous physical variables, such as object positions, orientations, and attributes (e.g., shape, size, and mass). These variables create an infinite set of possible goal configurations \(s^* \in \mathcal{S}\), making the mapping from state-action pairs to rewards inherently complex and highly dependent on the agents' coordination to address physical constraints. 
%
For instance, when two agents jointly move an object, the physical attributes of an object can influence the required actions for successful transitions.
Objects with irregular shapes require agents to coordinate their grips at specific parts.
Heavier objects demand synchronized forces of two agents.
%The state transition function \(\calP\) and the reward \(r\) are therefore contingent on these factors.
Considering the physical constraints \(\Gamma(s_t, a_t)\) that apply to the current state-action pair, the transition function is constrained as follows:
%Consider the following example where \(w\) represents the object’s weight, and \(a_t^i\) and \(a_t^j\) are the forces applied by the human and AI agents, respectively:
\[\small
\calP(s_{t+1} \mid s_t, a_t) = 
\begin{cases} 
1, & \text{if } \Gamma(s_t, a_t)\text{ satisfies } \text{(transition to $s_{t+1}$)} \\
0, & \text{otherwise (stay in } s_t\text{)}
\end{cases}
\]
\vspace{-1.0em}

These constraints create narrow transitions, similar to prior studies about motion planning~\citep{hsu2003bridge,saha2005finding,szkandera2020narrow}, and can further affect the agents' collaboration strategies. 
For example, in scenarios where the agents need to move a rectangular sofa through a narrow doorway, the agents need to grasp the shorter sides of the sofa and coordinate their moves to ensure they can fit through the entrance without collision.
%In this paper, we study human-AI collaboration under the challenges of continuous state space and constrained transitions introduced by physical embodiments and environments.
% \vspace{-0.6em}
\begin{figure}[!ht] %[!h]
    \centering
    \includegraphics[width=1.0\linewidth]{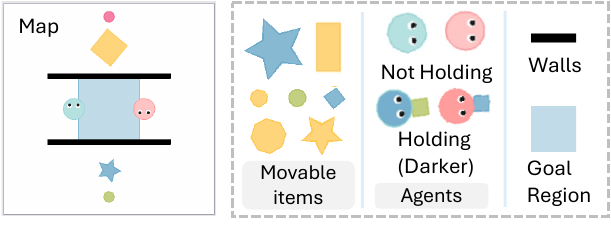}
    % \vspace{-0.8em}
    \caption{\textit{Moving Out} requires two agents to collaboratively move objects to the goal regions. The environment includes movable objects with varying shapes and sizes.
    % An agent can move a small item quickly. As the object sizes increase, the agent needs the other's help to move the object.
    %three object types: small objects, which can be moved quickly by one agent; middle-sized objects, which one agent can move slowly but two agents can move faster; and large objects, which require both agents to carry together.
    }
    \label{fig:introduction}
    \vspace{-1.5em}
\end{figure}

\section{Moving Out Benchmark}
%\vspace{-1em}
\subsection{Environment}
%\vspace{-6pt}
To test how physical environments can affect human-AI collaboration, we need an environment that follows physics.
We build \textit{Moving Out} on top of a 2D physics engine~\cite{Blomqvist_Pymunk_2025} with realistic rigid-body simulation.
%a single-agent environment Magical~\citep{toyer2020magical,zakka2021xirl} where agents and objects are physical bodies moving in a 2D physics simulation. 
Similar 2D physics engines have also been adopted in recent works studying physical reasoning and embodied AI~\citep{morlanscausal,ICLR2024_78834433,liu2024physgen}.
As shown in Fig.~\ref{fig:introduction}, each agent can maneuver freely in Moving Out and move objects with varying degrees of difficulty depending on the \updatetext{physical attributes of the object}.
The goal is to move all objects to the goal regions.
This design emphasizes flexibility, allowing agents to act independently while creating scenarios where collaboration is necessary for efficient task completion.

\begin{figure}[!ht] %[!h]
    \centering
    % \vspace{-1em}
    \includegraphics[width=1.0\linewidth]{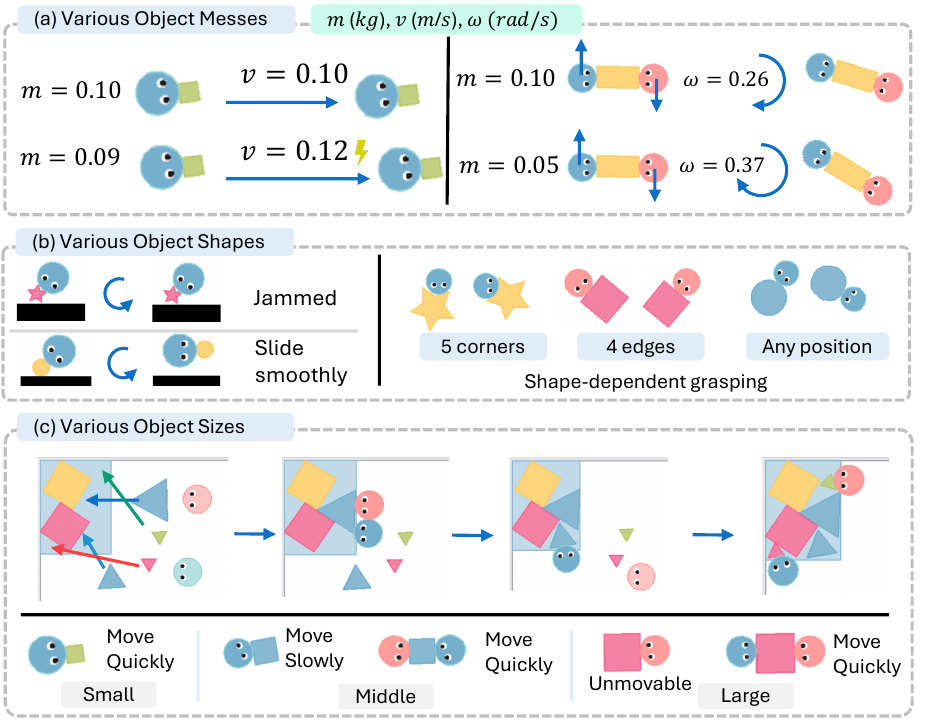}
    \vspace{-1em}
    \caption{\updatetext{Examples (not exhaustive) illustrating how different physical attributes affect task execution.
(a) Object mass influences the translational speed and angular velocity during transport.
(b) Object shape affects interactions with walls, such as sliding or getting stuck during contact, and different shapes require grasping different edges or corners.
(c) Object size affects the arrangement of items needed to successfully fit in the goal region.
Small objects are moved by one agent, medium objects move more efficiently with collaboration, and large objects require two agents.
}
    }
    \label{fig:physical_interaction}
    \vspace{-0.5em}
\end{figure}

% \vspace{-1em}
\subsubsection{Physical Variables}

%The environment includes movable items, walls, and goal regions. 
% \vspace{-10pt}

\textbf{Movable Items} are controlled by the following variables to introduce diverse physical interactions. \updatetext{Fig.~\ref{fig:physical_interaction} illustrates how different variables affect task execution.}
\vspace{-0.5em}
\begin{itemize}[noitemsep,topsep=0pt,parsep=0pt,partopsep=0pt,leftmargin=*]
    \item \textbf{Shapes} include stars, polygons, and circles, each requiring unique grabbing and rotation strategies.
    \item \textbf{Sizes} range from small to large, each has increasing difficulty in moving, can slow agent speed, and require more careful arrangement within the goal region.
    \item \textbf{Mass} is varied for different items. This influences an agent's moving speed during transportation.
\end{itemize}
% \vspace{-1em}
\textbf{Walls} introduce friction.
Agents that collide with walls experience reduced moving speed. 
%adding another layer of complexity. 

\textbf{Goal regions} are designated areas larger than the total size of items.
Agents must carefully arrange items to ensure all items can fit in the region, requiring precise spatial planning and coordination.
%\vspace{-1em}
%This combination of elements creates a dynamic and physically grounded environment, challenging agents to adapt their strategies, prioritize tasks, and collaborate effectively under varying constraints.

\begin{figure*}[!h]
    \centering
    \includegraphics[width=1.0\linewidth]{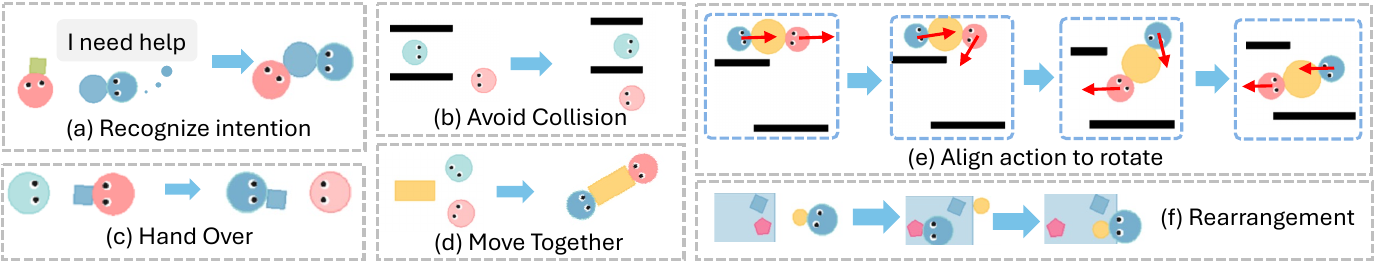}
    %\vspace{-0.6em}
    \caption{Diverse collaboration behaviors in \textit{Moving Out}, including (a) recognizing when help is needed, (b) avoiding collisions, (c) passing objects, (d) moving items together, (e) aligning actions, and (f) organizing objects in the goal region.}
    %\vspace{-0.5em}
    \label{fig:collaboration_modes}
\end{figure*}

\begin{figure}[!htp]
    \centering
    %\vspace{-1em}
    \includegraphics[width=1.0\linewidth]{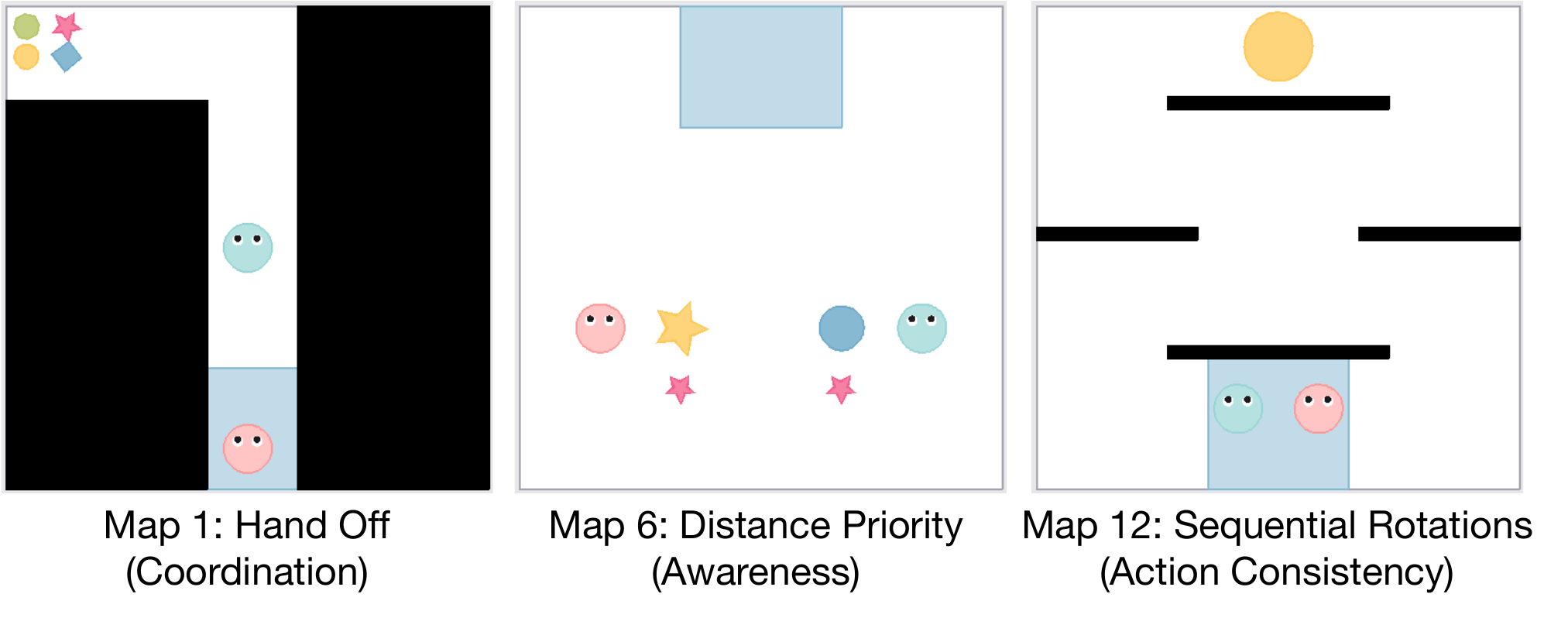}
    %\vspace{-1.5em}
    \caption{Example maps of different collaboration modes.
    %: coordination, awareness, and action consistency.
    }
    \label{fig:map_examples}
    \vspace{-0.5em}
\end{figure}

\subsubsection{Layout Types}

The physical variables introduce diverse collaborative behavior as shown in Fig.~\ref{fig:collaboration_modes}.
A successful collaboration usually requires a mixture of different behaviors.
To systematically understand the collaborative performance of AI agents, we designed 12 maps focusing on three collaboration modes.
See examples in Fig.~\ref{fig:map_examples} and Appendix \ref{appx:all_maps} for the full set.
%Each category reflects a specific aspect of human-robot collaboration and provides a structured framework for evaluating the agents' abilities in these areas.

\textbf{Coordination}~
The maps in this category only include small items, so each agent can complete the task independently.
However, narrow passages in the maps often block an agent’s path, requiring the partner to step aside or help pass the item.
For example, in Map 1 (Hand Off), the blue agent must pick up the item and, because of the narrow passage, pass it to the pink agent.
This setup enforces cooperation, as the task cannot be completed without coordination between the two agents.

\textbf{Awareness}~
The maps in this category do not have a clear optimal sequence for moving items, requiring agents to decide whether, when, and how to assist their partner for efficiency.
For instance, in Map 6 (Distance Priority), each agent starts near items and must decide whether to handle nearby items first or assist their partner.
These decisions become even more complex when collaborating with a human, as human behavior can vary significantly.
A human partner might wait for AI help with larger items, be passive, or focus on smaller tasks independently.
This variability demands that the AI agent dynamically adapts to the human's behavior.

\textbf{Action Consistency}~
This scenario requires agents to maintain consistent and synchronized actions over time, such as aligning their efforts to move and rotate large items.
The difficulty is aligning force directions and dynamically adjusting them to ensure efficient movement while navigating around tight spaces or obstacles.
For instance, in Map 12 (Sequential Rotations), two agents must collaboratively transport a large item through a series of narrow passages.
The agents must continuously synchronize their actions to adjust the item's angle, allowing it to fit through the openings.
Misalignment in their efforts could result in the item getting stuck or unnecessary movements that waste time and energy.

%\vspace{-0.5em}
\subsubsection{Abilities Required for Collaboration}
\updatetext{Achieving success in Moving Out requires a set of core abilities beyond individual action execution.
First, complex physical constraints and moving partners demand robust \emph{planning}, requiring agents to reason about spatial features.
Second, achieving the goal requires the agent to handle \emph{implicit communication}.
Agents must coordinate via action-based interactions whose effects are legible to their partners, allowing AI to interpret intent from behaviors and adapt its cooperative strategy.
Finally, collaboration requires \emph{anticipation} beyond simple reactivity.
The AI must predict the partner's needs and align its strategy to satisfy their expectations, ensuring timely assistance and fluid coordination.}

%\vspace{-1em}
%\vspace{-0.5em}
\subsection{\updatetext{Challenge Types}}
We design two \updatetext{challenges} to evaluate a model's ability to adapt to diverse human behaviors and to generalize to unseen physical constraints.

\textbf{\updatetext{Challenge} 1: Adapting to Diverse Human Behaviors}~
The first challenge of physically grounded human-AI collaboration arises from the continuous state-action space, which allows for a wide range of possible human behavior.
To test whether an agent can adapt to diverse human behavior, we fixed the configurations of the 12 maps and collected human-human collaboration data that demonstrate different ways to collaborate in the same maps.
These demonstrations represent a finite set of human behaviors.
In this challenge, we train a model on this dataset and test it with a new human or AI collaborator that represent different behaviors.
This setup assesses whether the model can adapt beyond the observed behaviors.
For an agent designed to assist humans effectively, learning to adapt from limited human demonstrations is crucial. 

\textit{\updatetext{Evaluation Protocol}}.~ 
%While we can evaluate agents with new human participants, a complementary protocol is needed to ensure reproducible results. 
Simply training on the full dataset requires us to recruit human participants to play against the model during every test and can lead to highly variable results.
%and evaluating with AI-AI collaboration only tests on seen behaviors and thus fails to measure adaptation.
To address the reproducibility issue, we split the dataset by participants into two disjoint splits, train separate AI agents on each split, and then evaluate them by letting the agents play with each other. 
This protocol provides a \updatetext{controllable and} reproducible proxy for testing generalization to unseen human behaviors.

\textbf{\updatetext{Challenge} 2: Generalizing to Unseen Physical Constraints}~
The second challenge arises from the physical constraints, which limit the possible transitions of given states.
To test whether the agent understands physical constraints, we randomized the physical attributes of objects in all maps to collect human-human interaction data that demonstrates how humans adapt to physical variables.
Again, we train a model with the dataset and evaluate it on maps with unseen object attributes.
To ensure the model learns the effects of physical constraints rather than memorizing them, we avoid having identical objects in the training and testing sets. 
In particular, the variation is defined compositionally over the object’s physical properties, ensuring that evaluation maps always include unseen combinations (e.g., a large star-shaped object is excluded from training whereas only small stars and large squares are present).
This forces the model to understand the impact of shape and type, and generalize across varying physical configurations.

%\vspace{-0.4em}
\textit{\updatetext{Evaluation Protocol}}.~  
Although evaluating directly with humans is possible, a more reproducible and efficient approach is to train agents on the full dataset and then test them via AI-AI play. 
Since evaluation maps contain object attributes not seen during training, this setup directly measures an agent’s ability to generalize to unseen physical constraints.

%\vspace{-0.3em}
\subsection{Dataset}

The data collection was approved by the Institutional Review Board (IRB).
Two human players control the agents with joysticks.
The game ran at 10Hz, and on average, each map took around 30 seconds (or 300 time steps) to transport all items. See Appendix~\ref{appx:data_coolection} for details.

For \updatetext{Challenge} 1, we recruited 36 college students as participants and collected over 1,000 human-human demonstrations (2,000 action sequences in total) across 12 maps.
This ensures that the dataset captures a wide range of human behaviors, providing sufficient diversity for training and testing the model's ability to generalize to unseen human strategies.
As shown in Table~\ref{table:task_1_diversity}, we compare the diversity of our dataset against datasets collected by RL agents or experts using Dynamic Time Warping (DTW; mean and variance), entropy, and coverage distance, showing ours has the best diversity. This demonstrates the effectiveness of recruiting diverse participants for data collection.
See Appendix~\ref{appx:task_1_diversity_analysis} for further details, including trajectory visualizations.

% \vspace{-0.8em}
\begin{table}[!htp]
\centering
\begin{adjustbox}{width=\linewidth}
\begin{tabular}{lcccc}
\toprule
Dataset & \begin{tabular}[c]{@{}c@{}}DTW \\ Mean ($\uparrow$)\end{tabular} 
        & \begin{tabular}[c]{@{}c@{}}DTW \\ Var ($\uparrow$)\end{tabular} 
        & \begin{tabular}[c]{@{}c@{}}Avg. Entropy \\ (KDE) ($\uparrow$)\end{tabular} 
        & \begin{tabular}[c]{@{}c@{}}Coverage \\ Distance (RBF) ($\uparrow$)\end{tabular} \\
\midrule
Moving Out \updatetext{Challenge} 1       & \textbf{7.013} & \textbf{6.065} & \textbf{0.888} & \textbf{0.899} \\
Expert dataset         & 4.642 & 3.029 & 0.757 & 0.744 \\
RL agent collected data & 4.358 & 2.499 & 0.683 & 0.626 \\
\bottomrule
\end{tabular}
\end{adjustbox}
\vspace{0.3em}
\caption{Dataset diversity across different data collection methods. Our dataset achieves consistently higher diversity compared to expert and RL agent datasets.
}
\label{table:task_1_diversity}
% \vspace{-0.5em}
\end{table}

% For \updatetext{challenge} 2, 
% %we emphasize the randomized attributes of objects rather than the variable behaviors.
% we used 4 expert players to collect 720 human-human demonstrations (1,440 action sequences in total), with 60 demonstrations per map.
% Each map includes randomized object physical attributes, where pose, mass, and size were varied by up to 10\%, while object types and shapes were randomized to be different from those used in evaluation.
% This setup allows us to assess the model's ability to generalize to unseen object attributes.
% Appendix~\ref{appx:challenge_2_map_example} shows examples of two maps.

\begin{figure}
    \centering
    \vspace{-1em}
    \includegraphics[width=1.0\linewidth]{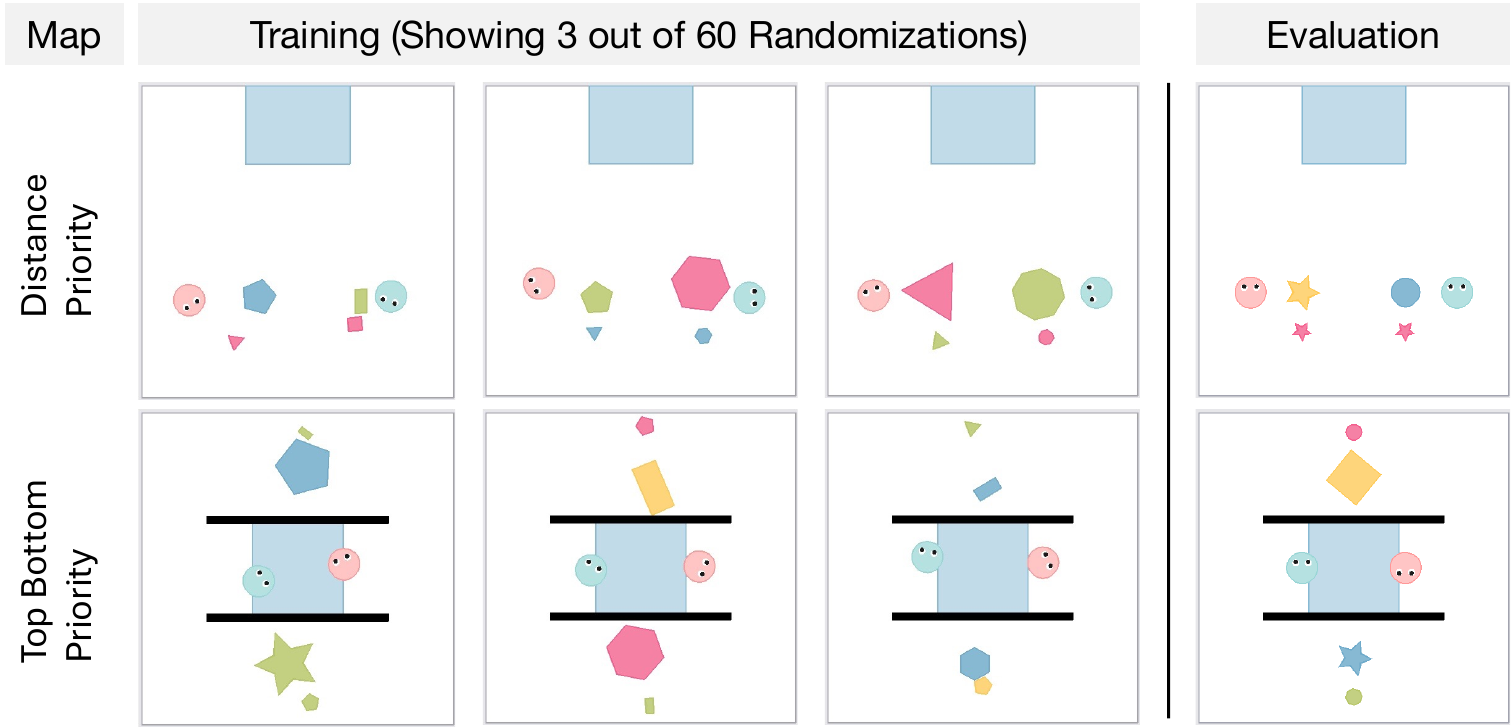}
    % \vspace{-1.5em}
    \caption{\updatetwotext{Examples of randomization in Challenge 2, illustrating generalization to unseen physical properties.}}
    \label{fig:visual_task2}
    \vspace{-1em}
\end{figure}
\updatetwotext{For Challenge 2, we emphasize the randomized properties of objects rather than the variable behaviors.
In this case, we used 4 expert players to collect 720 human-human demonstrations (1,440 action sequences in total), with 60 demonstrations per map.}
% Each map included randomized physical attributes, allowing us to evaluate the model's ability to generalize to unseen object attributes.
\updatetwotext{Each map included randomized object attributes, where pose, mass, and size were varied by up to 10\%, while object types and shapes were randomized to be different from those used in evaluation. This setup allows us to assess the model's ability to generalize to unseen object attributes. Fig.~\ref{fig:visual_task2} shows examples of two maps.}

\begin{figure*}[ht]
    \centering
    \includegraphics[width=1.0\linewidth]{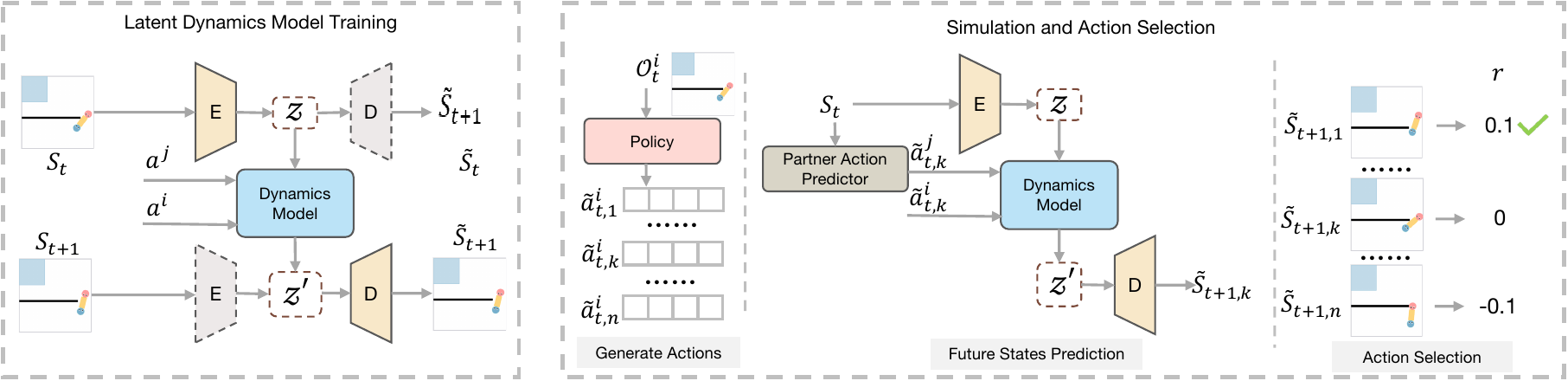}
    % \vspace{-0.6em}
    \caption{Overview of our Simulation and Action Selection components. \textbf{(Left)} The latent dynamics model that encodes the latent state from $t$ to $t+1$ to enable next state prediction. \textbf{(Right)} The action selection pipeline: The policy first generates candidate actions. The dynamics model then estimates the resulting future states, and finally, the best action is selected based on state evaluation.}
    % \vspace{-1.05em}
    \label{fig:method_faas}
\end{figure*}

% \vspace{-0.8em}
\section{BASS: Behavior Augmentation, Simulation, and Selection}

We develop BASS to address the increased number of configurations in continuous space and the outcome of actions in physical environments.
First, at training time, we augment the behavior data.
This helps the model adapt to diverse behaviors by exposing it to a broader range of possible interactions.
Second, we train a dynamics model to simulate the outcome of an action, allowing the agent to understand the impact of actions considering physical attributes and partner's actions.
At inference time, the model selects actions by evaluating the predicted states.

%vspace{-0.8em}
\subsection{Collaboration Behavior Augmentation}

\textbf{Behavior Augmentation}~
Trajectory augmentation has been used in single-agent settings~\citep{kim2024stitching,sussex2018stitched}.
Extending this idea to multi-agent raises new challenges: naively altering one agent’s behavior can break the consistency required for cooperation, since both agents must pursue aligned goals for the trajectory to be valid.
We adopt two strategies to generate valid augmentation.

First, we perturb partner poses with small noise, 
\(
\tilde{p}_{\text{partner}} = p_{\text{partner}} + \epsilon,  \epsilon \sim \mathcal{N}(0,\sigma^2),
\) 
while keeping other state variables unchanged, where \( p_{\text{partner}} \) is the original partner pose, \( \epsilon \) is Gaussian noise with mean \( 0 \) and variance \( \sigma^2 \), and \( \tilde{p}_{\text{partner}} \) is the perturbed pose used to generate new state variations. 
This generates new states that mimic natural variations in human motion and improve robustness to small deviations. See Appendix\ref{appx:pose_noise_anay} for the analysis of different noise scales.

Second, we augment the data by recombining sub-trajectories from two demonstrations.
In successful demonstrations, if agent~$i$'s sub-trajectories are the same, this indicates agent~$j$'s sub-trajectories are compatible with agent $i$'s, even if the ones from agent~$j$ are very different.
Taking this intuition, the key idea is to keep agent~$i$'s behavior fixed while swapping the partner's sub-trajectories.
We identify a segment of agent~$i$ in demonstration~A between timesteps $t_1$ and $t_2$, and another segment in demonstration~B between $t_3$ and $t_4$, where the agent begins and ends in nearly the same state. 
Because of the continuous state space, we treat two states as equivalent when the difference in the agent's pose is below a very small tolerance \(\epsilon_{\text{pose}}\), which is visually indistinguishable in practice.
Formally, this matching condition is expressed as \(s^i_{t_1} \approx \hat{s}^i_{t_3}\) and \(s^i_{t_2} \approx \hat{s}^i_{t_4}\) with \(t_2 > t_1\) and \(t_4 > t_3\).
Once these two segments match for agent~$i$, we keep agent~$i$'s motion unchanged and swap the corresponding partner~$j$’s sub-trajectories between the two demonstrations.
Further implementation details and visualization are provided in Appendix~\ref{appx:behavior_aug_details}.

\textbf{Validity of the Augmented States}~
We validate the generated sub-trajectories by checking whether the generated states remain within the valid state space and do not result in collisions or other inconsistencies. 
%However, we will show in the experiment section that, even without explicit validation, this augmentation strategy can improve performance.
As detailed in Appendix~\ref{appx:validation}, our recombination strategy is explicitly designed to ensure coherence, e.g., by swapping only sub-trajectories with identical start and end states, achieving a success rate above 99\% in producing physically valid trajectories.
% Even without additional validation, this augmentation strategy can improve performance, as shown in the experiments.

%\vspace{-0.8em}
\subsection{Simulation and Action Selection}

To understand the outcome of an action, in simulation environments, we can directly utilize the physics engine to simulate the outcome.
However, in the real world, a simulator is unavailable, a world model or next state predictor is required.
Fig.~\ref{fig:method_faas} shows the training and inference pipelines of our Simulation and Action Selection components.
%The policy generates multiple candidate actions for a given scenario and selects the most effective one by predicting future states and evaluating their outcomes.
%In simulation environments, this can be done by replicating the environment and simulating action outcomes.

\textbf{Next State Prediction}~ Our next-state predictor uses two autoencoders to estimate future states.
First, one autoencoder encodes the current state into the latent space. 
The dynamics model then takes this latent representation and the actions of both agents as input to predict the latent representation of the next state. 
Finally, this predicted latent representation is decoded by another autoencoder to reconstruct the next state.
Since the next state depends on the agent's own action and the partner's action, we use a partner action predictor to \updatetext{explicitly anticipate} the partner's action based on the current state.
Practically, the partner's predictor can share the same architecture as the agent's policy or directly use the agent's own policy by swapping its state with the partner's state to predict the partner’s action.
The dynamics model predicts the future state as:
\(
z_{t+1} = f(z_t, a_t, a_t^{(p)}),
\)
where \(z_t\) and \(z_{t+1}\) represent the latent spaces of the current and future states, \(a_t\) is the agent's action, \(a_t^{(p)}\) is the inferred partner's action, and \(f\) is the dynamics model.

\textbf{Action Selection}~
Our policy and partner action predictor both require strong multi-modal modeling capacity to generate diverse action candidates, which forms the basis for action selection. 
Once the next state is predicted, the reward for each action is computed based on the total distance of all objects to the goal region.
We use Normalized Final Distance (NFD) as defined in Sec.~\ref{sec:experiment}, but other metrics that measure partial progress of map completion also suffice.
We then select the action with the highest reward as the optimal action:
\(
a^* = \arg\max_{a_i} r(a_i), i = 1, 2, \dots, n,
\)
where \(r(a_i)\) is the reward for action \(a_i\).
This approach enables the model to choose the most effective action, even in real-world scenarios without access to a simulator.
A comparison of NFD against alternative objectives is provided in Appendix~\ref{appx:choose_NFD}.

% \vspace{-1em}
\section{Experiments}
\label{sec:experiment}
We aim to answer the following questions:
% \textbf{(RQ1)} How well do existing methods and BASS perform in physically grounded settings?
\textbf{(RQ1)} Does BASS adapt to unseen human behaviors with limited performance degradation?
\textbf{(RQ2)} Does BASS generalize to unseen physical constraints?
\textbf{(RQ3)} Does the multi-agent design in BASS effectively consider the partner's behavior?
% that does not use partner information
\textbf{(RQ4)} Does BASS work more effectively with humans in physically grounded collaboration?
\textbf{(RQ5)} What failure patterns do baselines and BASS exhibit?
To answer these questions, we evaluate all methods on two Moving Out \updatetext{challenges} and conduct a human study for further validation. For AI–AI collaboration, results are averaged over 20 runs.

% \vspace{-0.8em}
\subsection{Settings}
\textbf{Baselines}~
We compare BASS against these behavior cloning and RL baselines to predict actions:
\begin{itemize}[noitemsep,topsep=0pt,parsep=0pt,partopsep=0pt,leftmargin=*]
    \item \textbf{MLP} is a common behavior cloning baseline.
    \item \textbf{GRU} captures temporal dependencies of state and actions using recurrent connections.
    \item \textbf{Diffusion Policy (DP)}~\citep{chi2023diffusion} captures multimodal distribution and has demonstrated strong performance across various tasks.
    \item \textbf{MAPPO}~\citep{yu2022surprising} is a commonly used multi-agent RL method. It has demonstrated strong performance in cooperative games. See Appendix~\ref{appx:mappo_setting} for training details.
\end{itemize}

\textbf{BASS Implementation}~ 
BASS builds on the same diffusion policy backbone used in our baselines, serving as both the base policy and the partner action predictor because of its strong multi-modal modeling capacity.
We choose variational autoencoder (VAE)~\cite{kingma2013auto} as our encoder and decoder model.
The VAE and dynamics models are implemented as MLPs and co-trained.
For action selection, the policy and partner predictor independently sample 4 action candidates each. 
While increasing the number of samples could further improve accuracy, collaboration requires real-time inference; sampling four candidates ensures inference can be performed at 10Hz. 
Ablation studies on the sampling strategy, analyses of individual modules are provided in Appendix~\ref{appx:bass_sampling},~\ref{appx:bass_details}, ~\ref{app:bass_module_analysis}.

\textbf{Evaluation Metrics}~
We measure the success of collaboration using the following metrics:
(1) Task Completion Rate (TCR) for successful item delivery;
(2) Normalized Final Distance (NFD) for the distances between objects and the target, measuring partial progress;
(3) Waiting Time (WT) for the amount of time an agent waits for another agent to assist with large items;
and 4) Action Consistency (AC) for the degree of force alignment when moving items jointly, indicating coordination efficiency.
Detailed definitions of these metrics are in Appendix~\ref{appx:evaluation_metrics}.

\textbf{Human Subject Study}~
Our study was approved by the IRB.
We conducted a human subject study with 32 participants to evaluate BASS against the DP baseline. 
Each participant played 12 maps in total, cooperating with each model in two rounds per map.
After completing the first round, the participant and model switched to control the other agent. 
Upon finishing all maps, participants were given a questionnaire for subjective feedback.
See details in Appendix~\ref{appx:human_study}.

% \vspace{-5pt}
% \begin{figure*}[!ht]
%     \centering
%     \includegraphics[width=1.0\linewidth]{imgs/t2_bar_results.jpg}
%     \vspace{-1.5em}
%     \caption{BASS outperforms baselines on Task 1 and Task 2, against AI (solid) and humans (striped).}
%     \vspace{-2.5ex}
%     \label{fig:results}
% \end{figure*}
% \vspace{-0.5em}

% \begin{wraptable}{r}{0.6\linewidth} 
% \vspace{-1em}
\begin{table}[!htp]
\centering
% \centering
\begin{adjustbox}{width=1.0\linewidth}
\begin{tabular}{@{}llllll@{}}
\toprule
\begin{tabular}[c]{@{}l@{}}Evaluation\\ Protocol\end{tabular}               & Method  & TCR (↑)                                                               & NFD (↑)                                                              & WT (↓)                                                               & AC (↑)                                                               \\ \midrule
\multirow{5}{*}{\begin{tabular}[c]{@{}l@{}}Seen\\ Behaviors\end{tabular}}   & MLP     & 0.2126                                                                & 0.2987                                                               & 0.4896                                                               & 0.8013                                                               \\
                                                                            & GRU     & 0.2369                                                                & 0.3011                                                               & 0.4975                                                               & 0.8151                                                               \\
                                                                            & MAPPO   & 0.1929                                                                & 0.3182                                                               & 0.5766                                                               & 0.8097                                                               \\
                                                                            & DP      & 0.3233                                                                & 0.5367                                                               & 0.3789                                                               & 0.8163                                                               \\
                                                                            & DP/BASS & \textbf{0.3503}                                                       & \textbf{0.5724}                                                      & \textbf{0.3598}                                                      & \textbf{0.8337}                                                      \\ \midrule
                                                                            & MLP     & \begin{tabular}[c]{@{}l@{}}0.1433 \\ (-32.61\%)\end{tabular}          & \begin{tabular}[c]{@{}l@{}}0.2413 \\ (-19.22\%)\end{tabular}         & \begin{tabular}[c]{@{}l@{}}0.5647 \\ (+15.33\%)\end{tabular}         & \begin{tabular}[c]{@{}l@{}}0.7729 \\ (-3.54\%)\end{tabular}          \\ \cmidrule(l){2-6} 
\multirow{3}{*}{\begin{tabular}[c]{@{}l@{}}Unseen\\ Behaviors\end{tabular}} & GRU     & \begin{tabular}[c]{@{}l@{}}0.1638 \\ (-30.87\%)\end{tabular}          & \begin{tabular}[c]{@{}l@{}}0.2453 \\ (-18.53\%)\end{tabular}         & \begin{tabular}[c]{@{}l@{}}0.5758 \\ (+15.74\%)\end{tabular}         & \begin{tabular}[c]{@{}l@{}}0.7830 \\ (-3.94\%)\end{tabular}          \\ \cmidrule(l){2-6} 
                                                                            & MAPPO   & \begin{tabular}[c]{@{}l@{}}0.1635 \\ (-15.19\%)\end{tabular}          & \begin{tabular}[c]{@{}l@{}}0.2808 \\ (-11.74\%)\end{tabular}         & \begin{tabular}[c]{@{}l@{}}0.6379 \\ (+10.64\%)\end{tabular}         & \begin{tabular}[c]{@{}l@{}}0.7858 \\ (-2.95\%)\end{tabular}          \\ \cmidrule(l){2-6} 
                                                                            & DP      & \begin{tabular}[c]{@{}l@{}}0.2563 \\ (-20.72\%)\end{tabular}          & \begin{tabular}[c]{@{}l@{}}0.4589 \\ (-14.50\%)\end{tabular}         & \begin{tabular}[c]{@{}l@{}}0.4249 \\ (+12.15\%)\end{tabular}         & \begin{tabular}[c]{@{}l@{}}0.7854 \\ (-3.78\%)\end{tabular}          \\ \cmidrule(l){2-6} 
                                                                            & DP/BASS & \textbf{\begin{tabular}[c]{@{}l@{}}0.3010 \\ (-14.07\%)\end{tabular}} & \textbf{\begin{tabular}[c]{@{}l@{}}0.5197 \\ (-9.22\%)\end{tabular}} & \textbf{\begin{tabular}[c]{@{}l@{}}0.3899 \\ (+8.37\%)\end{tabular}} & \textbf{\begin{tabular}[c]{@{}l@{}}0.8099 \\ (-2.86\%)\end{tabular}} \\ \midrule
\multirow{2}{*}{\begin{tabular}[c]{@{}l@{}}Play with\\ Human\end{tabular}}  & DP      & 0.3855                                                                & 0.5547                                                               & 0.4886                                                               & 0.8054                                                               \\
                                                                            & DP/BASS & \textbf{0.6512}                                                       & \textbf{0.7053}                                                      & \textbf{0.3364}                                                      & \textbf{0.9124}                                                      \\ \bottomrule
\end{tabular}
\end{adjustbox}
\vspace{0.5em}
\caption{Results of \updatetext{Challenge} 1 under seen and unseen human behaviors, and with real human partners.}
\label{table:task_1_results}
%\vspace{-1.5em}
% \end{wraptable}
\end{table}

% \vspace{-1em}
\subsection{Results}
\textbf{Collaboration with Unseen Behaviors (RQ1)}~
Table~\ref{table:task_1_results} reports results of \updatetext{Challenge} 1 under three protocols. 
In the seen setting, agents are trained and evaluated on the same dataset. 
Here, both DP and BASS outperform MLP, GRU, and MAPPO, with BASS achieving the best task completion (TCR, NFD). 
In the unseen setting, we split participants into disjoint sets as described in the evaluation protocol, train separate policies, and then evaluate them by playing across groups. 
All methods degrade when facing unseen behaviors, but BASS shows the least performance drop across TCR, NFD, WT, and AC, indicating stronger robustness.
See Appendix~\ref{app:task1_full_results} for the full table with standard error.

\begin{figure}[!htp]
    \centering
    %\vspace{-0.8em}
    \includegraphics[width=1.0\linewidth]{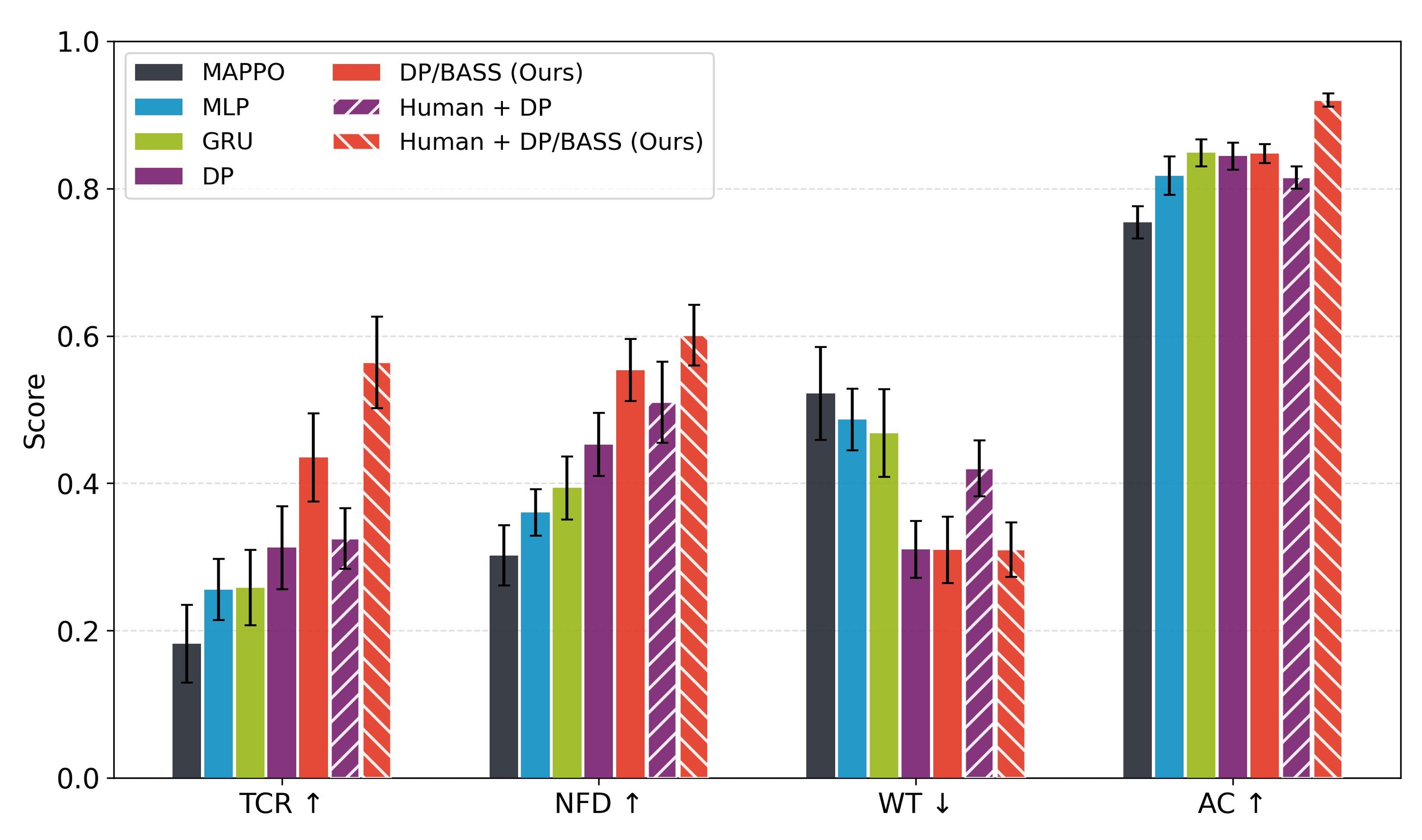}
    \vspace{-1em}
    \caption{Results of \updatetext{Challenge} 2 under unseen physical constraints.}
    \label{fig:results_t2}
    \vspace{-0.5em}
\end{figure}

\textbf{Collaboration under Unseen Physical Constraints (RQ2)}~
Fig.~\ref{fig:results_t2} shows that, while waiting time and action consistency are comparable across methods, BASS consistently outperforms baselines, particularly in TCR and NFD, which directly reflect task progress under new object properties. 
This suggests that evaluating candidate actions based on predicted future states helps the model better adapt to variations in size, mass, and shape. 
%While waiting time and action consistency are comparable across methods, BASS achieves better overall task completion under unseen physical constraints.

\paragraph{Effectiveness of the Multi-agent Design in BASS (RQ3)}~
To show the importance of modeling both agents, we compare BASS with a single-agent variant that ignores partner alignment during recombination and predicts only one agent's future state during action simulation.
The single-agent variant increases diversity but noticeably reduce collaboration performance.
For example, in \updatetext{Challenge}~1, TCR and NFD drop from \{0.403, 0.511\} in the full multi-agent version to \{0.368, 0.451\} when using single-agent recombination.
In \updatetext{Challenge}~2, the multi-agent action simulation achieves \{0.420, 0.554\}, whereas the single-agent variant reduces these to \{0.319, 0.458\}.
These results show that a multi-agent model structure is necessary for generating valid augmented trajectories and selecting effective actions.
Full results and tables are provided in Appendix~\ref{appx:comparsion_single_agent}.

\textbf{Collaboration with Humans (RQ4)}~
Tab.~\ref{table:task_1_results} and Fig.~\ref{fig:results_t2} show the results with humans. 
In both \updatetext{challenges}, BASS significantly improved task completion rates (TCR and NFD) compared to the DP.
%, which showed smaller improvements. 
This indicates that BASS adapts better to human behavior.
%, enhancing overall performance. 
For wait time, DP increased when playing with humans, suggesting it struggles with different humans, despite DP capturing multimodal distributions. 
BASS reduced wait time, demonstrating its ability to adapt to diverse behaviors. 
For action consistency, DP performed worse because it cannot handle differences between the evaluation and training data. 
BASS augmented diverse collaborative behaviors during training and selected the best actions for interacting with humans, resulting in better consistency.

\begin{figure}[!htp]
    \centering
    \vspace{-0.5em}
    \includegraphics[width=1.0\linewidth]{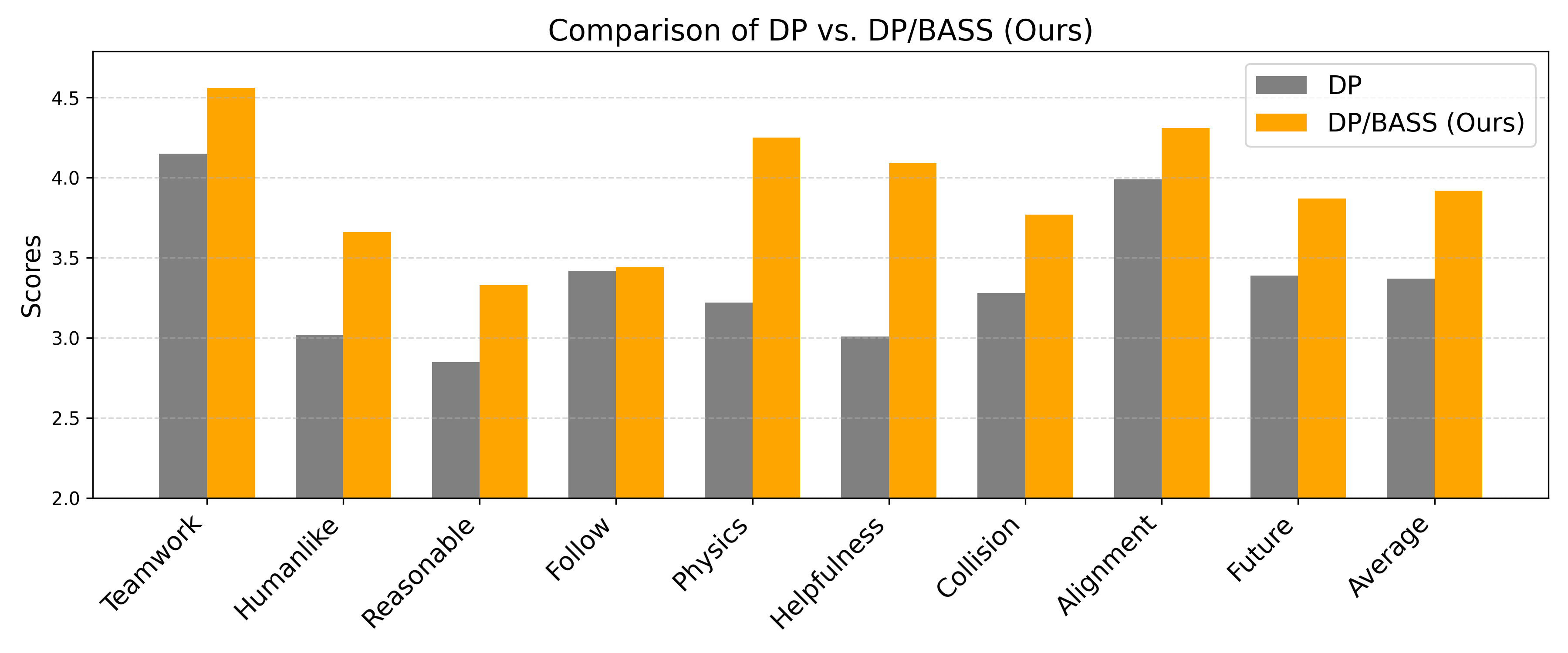}
    \vspace{-1em}
    \caption{User survey results in a 7-point Likert scale}
    \label{fig:human_study_results}
    \vspace{-0.5em}
\end{figure}
\textbf{Human Feedback (RQ4)}~
Fig.~\ref{fig:human_study_results} summarizes post-experiment survey results from humans.
%, with detailed questions available in Appx~\ref{appx:questionnaire}.
We compare BASS with DP.
The results show that BASS significantly outperformed DP in the Helpfulness category, indicating that BASS is better at consciously assisting others. 
Additionally, BASS demonstrated a better understanding of physics, suggesting that our next state predictor effectively comprehends and evaluates different actions to choose the best ones. 
Independent t-tests revealed that these differences are statistically significant ($p = 0.017$).

% \begin{wrapfigure}{r}{0.48\textwidth}
%         \centering
%         \begin{scalebox}{0.75}{
%         \begin{tabular}{ccccc}
%         & \multicolumn{2}{c}{NFD↑} & \multicolumn{2}{c}{Prediction Accuracy} \\ \cline{2-5} 
%         & Task 1      & Task 2     & Task 1             & Task 2             \\ \hline
%         DP + BASS                                                               & 0.5733      & 0.5535     & 0.6250             & 0.4870             \\ \hline
%         \begin{tabular}[c]{@{}c@{}}DP + BASS \\ w/Oracle Simulator\end{tabular} & 0.5875      & 0.6209     & N/A                  & N/A                 
%         \end{tabular}
%         }
%         \caption{Performance of different simulation strategies. The oracle simulator serves as the upper bound for our method.}
%          \label{tab:simulator}
%         \end{scalebox}
% \end{wrapfigure}

\begin{figure}[!htp]
    \centering
    \vspace{-0.5em}
    \includegraphics[width=1.0\linewidth]{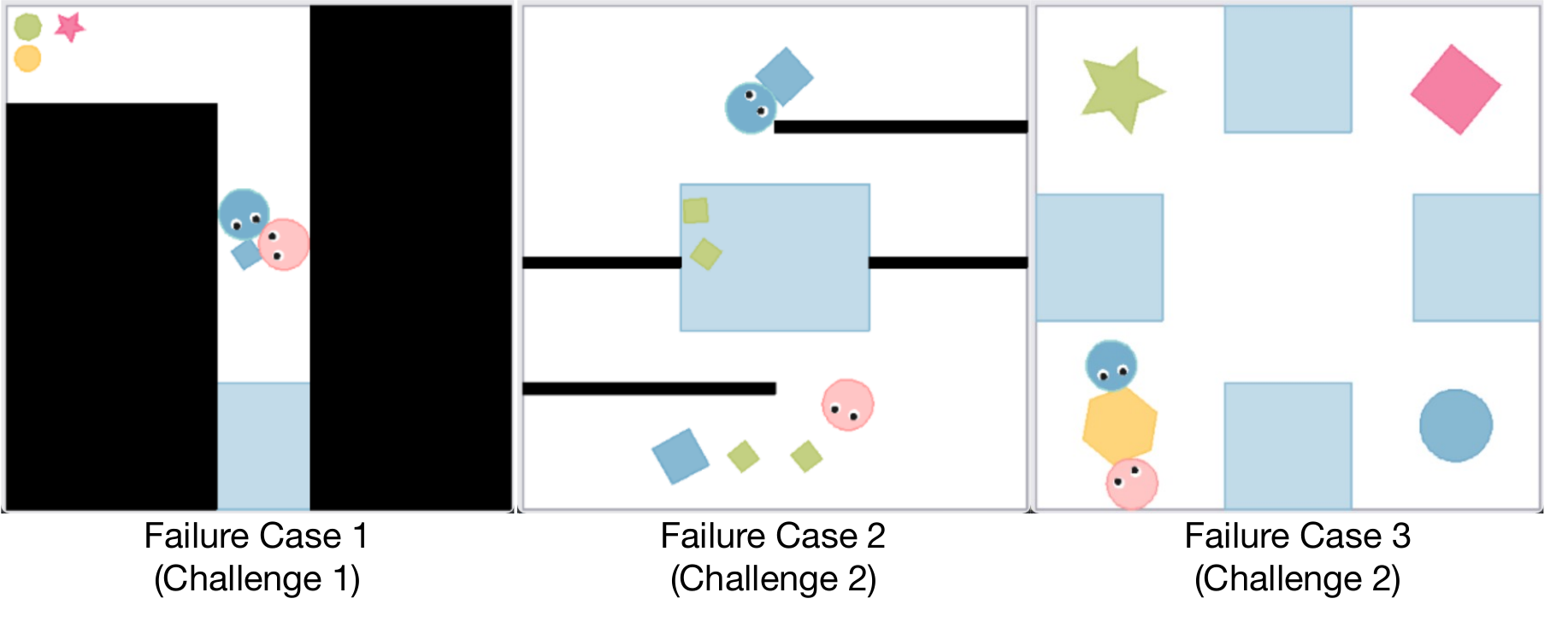}
    \vspace{-1em}
    \caption{Failure cases: 1) Failing to release items during handover, 2) Not responding when assistance is needed, and 3) Inability to grasp large items upon approach.}
    \label{fig:failure_cases}
    \vspace{-0.5em}
\end{figure}

\textbf{Failure case study (RQ5)}~
Fig.~\ref{fig:failure_cases} shows examples of common failure cases from DP.
In \updatetext{Challenge} 1, as illustrated in failure case 1, many participants reported that the AI agent frequently holds an item without passing it, resulting in frequent collisions. 
Additionally, participants noted that the AI agent often failed to come to assist, as shown in failure case 2, where a human agent (blue) was slowly pulling an item, but the AI agent (pink) instead went to grasp other smaller objects. 
These issues show the model's limited ability to adapt to diverse behaviors.
%, making it difficult to respond appropriately to actions that were not present in the training dataset. 
In \updatetext{Challenge} 2, most participants pointed out failure case 3, where the AI agent reached the target item but was unable to successfully grasp it.
This indicates that the model struggles when encountering objects not in the training data.
In contrast, BASS shows fewer reported failure cases than DP.
Manual inspection revealed that BASS reduced the occurrence rates of the three failure types from \{0.797, 0.688, 0.906\} in DP to \{0.343, 0.563, 0.484\}.
However, effectively addressing these failures remains a substantial challenge for future research.

% In contrast, BASS shows fewer reported failure cases than DP, but effectively addressing these failures remains a substantial challenge for future research.

% Overall, DP and BASS both have these failure cases.
% However, compared to DP, BASS received fewer reports about these failures, demonstrating its adaptability and overall performance improvement. %This suggests that our BASS approach enhances the model’s ability to effectively interact and collaborate with human partners.

% \vspace{-1em}
\section{Conclusion}
% \vspace{-1em}

We introduce \textit{Moving Out}, a physically grounded human-AI collaboration benchmark that features
a continuous state-action space and dynamic object interactions.
We created two challenges and collected human-human collaboration data to enable future model development.
Our results show that much remains to be done with existing models to effectively collaborate with humans in physical environments.
Our proposed method, BASS, takes the first step to improve models’ adaptability to diverse human behaviors and physical constraints. 
Future work includes strengthening the theoretical understanding of human–AI collaboration and the BASS framework, extending the benchmark toward richer and more complex cooperative scenarios, and adding explicit communication on top of the implicit communication already present in our current environment.
These directions will enhance both the practical and theoretical aspects of physically grounded human–AI collaboration.

% \clearpage

\section*{Acknowledgment}
This work was supported by Toyota Research Institute and NSF CMMI-2443076.
We acknowledge Research Computing at the University of Virginia for providing the computational resources and technical support that made the results in this
work possible.

\section*{Impact Statement}

The integration of AI into real-world applications requires effective human-AI collaboration, yet existing research environments often fail to capture the complexities of physical interactions. 
Moving Out provides a physically grounded framework to study and improve human-AI cooperation in dynamic settings, making it a valuable tool for advancing AI’s role in real-world applications such as assistive robotics, industrial automation, and interactive AI systems. 
By enabling AI agents to adapt to diverse human behaviors and physical constraints, our environment contributes to the development of more adaptive, efficient, and human-compatible AI systems.

Beyond the environment, our proposed methods, Behavior Augmentation and Simulation and Action Selection, enhance AI’s ability to understand human intentions and collaborate more effectively. 
These techniques can be applied broadly to assistive, multi-agent, and interactive systems, ensuring AI agents support human partners more intuitively and reliably. 
While increasing AI’s autonomy in collaborative settings raises ethical considerations regarding trust, safety, and human control, our work provides a structured foundation to study and address these challenges, fostering responsible AI development that aligns with human needs.


\bibliography{ref}
\bibliographystyle{icml2026}

%%%%%%%%%%%%%%%%%%%%%%%%%%%%%%%%%%%%%%%%%%%%%%%%%%%%%%%%%%%%%%%%%%%%%%%%%%%%%%%
%%%%%%%%%%%%%%%%%%%%%%%%%%%%%%%%%%%%%%%%%%%%%%%%%%%%%%%%%%%%%%%%%%%%%%%%%%%%%%%
% APPENDIX
%%%%%%%%%%%%%%%%%%%%%%%%%%%%%%%%%%%%%%%%%%%%%%%%%%%%%%%%%%%%%%%%%%%%%%%%%%%%%%%
%%%%%%%%%%%%%%%%%%%%%%%%%%%%%%%%%%%%%%%%%%%%%%%%%%%%%%%%%%%%%%%%%%%%%%%%%%%%%%%
\newpage
\appendix
\onecolumn

% \section{Impact Statement}
% \label{appx:impact_statement}
% The integration of AI into real-world applications requires effective human-AI collaboration, yet existing research environments often fail to capture the complexities of physical interactions. 
% Moving Out provides a physically grounded framework to study and improve human-AI cooperation in dynamic settings, making it a valuable tool for advancing AI’s role in real-world applications such as assistive robotics, industrial automation, and interactive AI systems. 
% By enabling AI agents to adapt to diverse human behaviors and physical constraints, our environment contributes to the development of more adaptive, efficient, and human-compatible AI systems.

% Beyond the environment, our proposed methods—Behavior Augmentation and Simulation and Action Selection—enhance AI’s ability to understand human intentions and collaborate more effectively. 
% These techniques can be applied broadly to assistive AI, multi-agent systems, and interactive AI models, ensuring AI agents support human partners more intuitively and reliably. 
% While increasing AI’s autonomy in collaborative settings raises ethical considerations regarding trust, safety, and human control, our work provides a structured foundation to study and address these challenges, fostering responsible AI development that aligns with human needs.

\newpage
\section{Comparison with Other Environments}
\label{appx:comparsion}
\begin{table}[h]
\begin{adjustbox}{width=1.0\linewidth}
\begin{tabular}{|c|c|c|c|c}
\hline
\textbf{Environment} & \textbf{State/Action} & \textbf{\begin{tabular}[c]{@{}c@{}}Physics\\ -based\end{tabular}} & \textbf{Constraints}                                                                                                                      & \textbf{Collaboration Behaviors}                                                                                                     \\ \hline
Overcooked-AI        & Discrete              & No                                                                & \begin{tabular}[c]{@{}c@{}}Items placed only \\ in specific locations\end{tabular}                                                        & \begin{tabular}[c]{@{}c@{}}Passing items, dividing tasks, \\ and collision avoidance\end{tabular}                                    \\ \hline
Watch and Help       & Discrete              & Yes                                                               & \begin{tabular}[c]{@{}c@{}}Partial observability, \\ diverse objects and goals\end{tabular}                                               & \begin{tabular}[c]{@{}c@{}}Goal inference, \\ cooperative help\end{tabular}                                                          \\ \hline
Smart Help           & Discrete              & Yes                                                               & \begin{tabular}[c]{@{}c@{}}Capability limits \\ (weight, height, open/close/toggle), \\ partial observability\end{tabular}                & \begin{tabular}[c]{@{}c@{}}Awareness: \\ Goal + capability inference, bottleneck \\ help, avoid unnecessary takeover\end{tabular}    \\ \hline
Table Carrying       & Continuous            & No                                                                & \begin{tabular}[c]{@{}c@{}}No physical feedback,\\ task ends upon collision\end{tabular}                                                  & \begin{tabular}[c]{@{}c@{}}Joint carrying \\ (i.e., action consistency)\end{tabular}                                                 \\ \hline
\updatetext{HumanThor}            & Continuous            & Yes                                                               & \begin{tabular}[c]{@{}c@{}}Predefined object interactions, \\ scripted rules, \\ no continuous physical coupling\end{tabular}             & \begin{tabular}[c]{@{}c@{}}Division of labor in navigation \\ and rearrangement \\ (no joint manipulation)\end{tabular}              \\ \hline
\updatetext{Habitat}              & Continuous            & Yes                                                               & \begin{tabular}[c]{@{}c@{}}Predefined or pre-trained \\ low-level skills,\end{tabular}                                                    & \begin{tabular}[c]{@{}c@{}}Task-level coordination and division \\ of labor  (navigation, rearrangement)\end{tabular}                \\ \hline
Moving Out           & Continuous            & Yes                                                               & \begin{tabular}[c]{@{}c@{}}Realistic physics, friction, \\ collision feedback, \\ diverse items with \\ physical properties.\end{tabular} & \begin{tabular}[c]{@{}c@{}}Coordination, Awareness \\ of needing help, \\ joint carrying \\ (i.e., action consistency).\end{tabular} \\ \hline
\end{tabular}
\end{adjustbox}
\end{table}

\begin{table}[h]
\begin{adjustbox}{width=1.0\linewidth}
\begin{tabular}{c|c|c|c|}
\hline
\textbf{Metrics}                                                                                                                     & \textbf{Pros}                                                                                                                             & \textbf{Cons}                                                                                                                          & \textbf{Human Data} \\ \hline
\begin{tabular}[c]{@{}c@{}}Number of cooked \\ onions in a limited time\end{tabular}                                                 & \begin{tabular}[c]{@{}c@{}}Small state/action space, \\ fast training, \\ human data available\end{tabular}                               & \begin{tabular}[c]{@{}c@{}}Limited behavior variety, \\ simple tasks\end{tabular}                                                      & Yes                 \\ \hline
\begin{tabular}[c]{@{}c@{}}Success Rate,\\ speedup,\\ cumulative reward\end{tabular}                                                 & \begin{tabular}[c]{@{}c@{}}3D environment,\\ diverse household tasks\end{tabular}                                                         & \begin{tabular}[c]{@{}c@{}}No physical variations,\\ high computational cost\end{tabular}                                              & Synthesized         \\ \hline
\begin{tabular}[c]{@{}c@{}}Success Rate (Goal-conditioned),\\ Helping Necessity/Rate,\\ Episode/Success-weighted Length\end{tabular} & \begin{tabular}[c]{@{}c@{}}3D physics, \\ diverse tasks\end{tabular}                                                                      & \begin{tabular}[c]{@{}c@{}}Predefined actions,\\ high computational cost\end{tabular}                                                  & No                  \\ \hline
\begin{tabular}[c]{@{}c@{}}Success rate, \\ Completion time\end{tabular}                                                             & Continuous actions                                                                                                                        & \begin{tabular}[c]{@{}c@{}}No physics in interactions, \\ single task, no dataset\end{tabular}                                         & No                  \\ \hline
\begin{tabular}[c]{@{}c@{}}Success Rate, \\ Navigation Efficiency, \\ Object Placement Accuracy\end{tabular}                         & VR Integration                                                                                                                           & \begin{tabular}[c]{@{}c@{}}Rule-based agents only, \\ no learning or evaluation of \\ human-AI collaboration, no datasets\end{tabular} & No                  \\ \hline
\begin{tabular}[c]{@{}c@{}}Success rate of navigation \\ and rearrangement\end{tabular}                                              & Different embodiments,                                                                                                                    & \begin{tabular}[c]{@{}c@{}}Manipulation relies on predefined \\ or pre-trained skills, \\ no collaboration datasets\end{tabular}       & No                  \\ \hline
\begin{tabular}[c]{@{}c@{}}Task Completion Rate, \\ Normalized Final Distance, \\ Waiting Time,  Action Consistency\end{tabular}     & \begin{tabular}[c]{@{}c@{}}Realistic physics, \\ multiple collaboration modes, \\ physics feedback,  human dataset available\end{tabular} & \begin{tabular}[c]{@{}c@{}}Requires high-frequency actions \\ for smooth collaboration.\end{tabular}                                    & Yes                 \\ \hline
\end{tabular}
\end{adjustbox}
% \vspace{-2em}
\caption{Comparison between Moving Out, Overcooked-AI, and Table Carrying. Overall, Moving Out offers more diverse collaboration modes and physical constraints due to its physics-based environment.}
\vspace{-1em}
\end{table}

\section{\updatetext{Challenge} 1 Data Diversity Analysis}
\label{appx:task_1_diversity_analysis}
\subsection{Visualization of Dataset in \updatetext{Challenge} 1}
\begin{figure}[!h]
    \centering
    \includegraphics[width=0.7\linewidth]{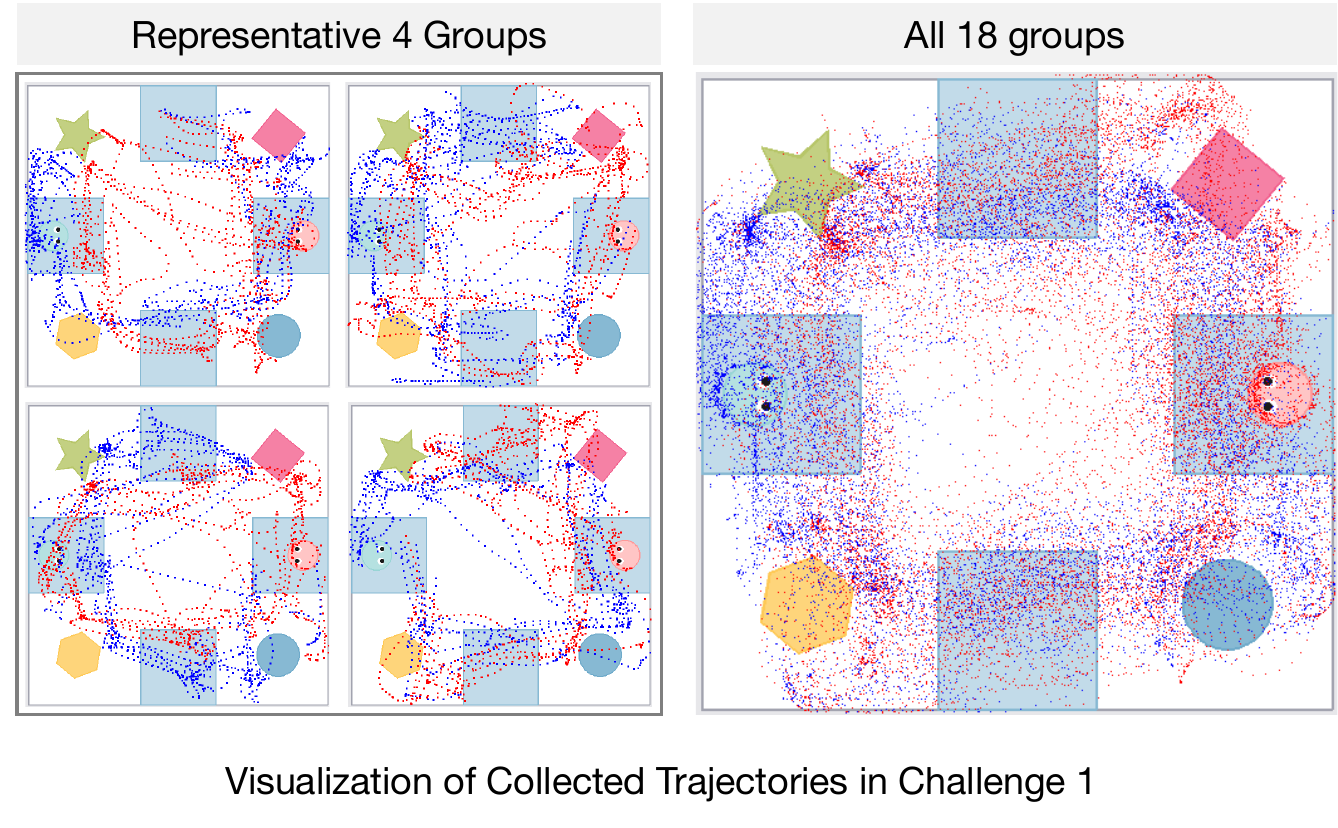}
    \vspace{-1em}
    \caption{Visualization of data collected in \updatetext{Challenge} 1 from four groups (each with two players) and from the complete dataset. Blue dots denote the positions of the blue agent and red dots denote the positions of the red agent. The visualizations show clear differences across groups. At the aggregated level, the dataset captures both human behavioral preferences (e.g., preferred paths and object-grasping locations) and broad state coverage.}
    \vspace{-1em}
    \label{fig:visual_task1}
\end{figure}
See Fig.~\ref{fig:visual_task1}.

\subsection{Evaluate the diversity of \updatetext{Challenge} 1}

To further quantify behavioral diversity in our datasets, we evaluated three distinct sources: the human-human dataset from Moving Out \updatetext{Challenge} 1, a dataset collected from four human experts, and trajectories generated by a trained MAPPO agent. We report diversity across both trajectory- and state-level dimensions. At the trajectory level, we use Dynamic Time Warping (DTW) to compute pairwise distances between all trajectories within a dataset; higher mean and variance indicate greater dissimilarity in path shapes. At the state level, we measure spatial coverage using two complementary metrics: Kernel Density Estimation (KDE) entropy over agent positions and an additional coverage distance metric inspired by~\citep{fu2023iteratively}, which computes the average pairwise distance between trajectories using an RBF kernel. Higher values for both metrics indicate broader exploration of the map. As shown in Table~\ref{table:task_1_diversity}, the \updatetext{Challenge} 1 dataset consistently achieves higher scores across all metrics, confirming that data aggregated from 36 human players exhibits substantially greater behavioral diversity compared to expert demonstrations and RL-generated trajectories. This diversity provides a rich foundation for training adaptive collaboration policies and benchmarking generalization.

% \section{Randomization Examples in Challenge 2}
% \label{appx:challenge_2_map_example}
% \begin{figure}[!htp]
%     \centering
%     \vspace{-1em}
%     \includegraphics[width=0.8\linewidth]{imgs/visual_task_2.pdf}
%     % \vspace{-1.5em}
%     \caption{Randomization examples in \updatetext{challenge} 2, illustrating generalization to unseen physical properties.}
%     \label{fig:visual_task2}
%     \vspace{-1em}
% \end{figure}

\section{Comparison with Oracle Simulation}
\label{appx:comparsion_with_oracle_simulation}
\begin{table}[!h]
        \centering
        \begin{scalebox}{1.0}{
        \begin{tabular}{ccccc}
        & \multicolumn{2}{c}{NFD↑} & \multicolumn{2}{c}{Prediction Accuracy} \\ \cline{2-5} 
        & \updatetext{Challenge} 1      & \updatetext{Challenge} 2     & \updatetext{Challenge} 1             & \updatetext{Challenge} 2             \\ \hline
        DP + BASS                                                               & 0.5733      & 0.5535     & 0.6250             & 0.4870             \\ \hline
        \begin{tabular}[c]{@{}c@{}}DP + BASS \\ w/Oracle Simulator\end{tabular} & 0.5875      & 0.6209     & N/A                  & N/A                 
        \end{tabular}
        }
        \caption{Performance of different simulation strategies. The oracle simulator serves as the upper bound for our method.}
         \label{tab:simulator}
        \end{scalebox}
\end{table}
We compare the task completion (NFD) and prediction accuracy of actions against the oracle simulator (i.e., the 2D physics engine) in Table~\ref{tab:simulator}.
We compute the prediction accuracy by comparing the actions selected using our next state predictor versus the actions selected using the oracle simulator.
The oracle simulator serves as the upper bound for our action selection method since it provides the ground-truth next states.
%This also allows us to evaluate the accuracy of our action selection. 
We observe that our model achieves higher accuracy in \updatetext{Challenge} 1, with results that are closer to those of the oracle simulator.
This is because \updatetext{Challenge} 1 uses a fixed map, while \updatetext{Challenge} 2 trains on randomized states. 
%So in task 1, the future state predictor can be trained with higher accuracy, leading to better performance.

\section{Ablation Study}
\label{appx:ablation}

\textbf{Ablations}~
Table \ref{tab:ablation} shows the ablation of each component.
%This confirms the improvements of each component.
%By examining the table, we can see that incorporating Behavior or Simulation and Action Selection individually yields consistent improvements over the baseline GRU and Diffusion Policy (DP) models on both Task 1 and Task 2. 
Adding augmentation and simulation components improves task completion TCR and NFD compared to their base models. 
When using all components (full BASS), they achieve the highest overall performance in most cases.
%This pattern suggests that behavior augmentation and action selection each provide complementary benefits, and their combined use offers the largest performance gains.

\begin{table}[h]
    \centering
    \begin{scalebox}{1.0}{
    \begin{tabular}{lcccc}
                      & \multicolumn{2}{c}{\updatetext{Challenge} 1}        & \multicolumn{2}{c}{\updatetext{Challenge} 2}        \\ \hline
Methods               & TCR↑            & NFD↑            & TCR↑            & NFD↑            \\
GRU                   & 0.3070          & 0.3674          & 0.2582          & 0.3935          \\
+ BASS w/o Simulation & 0.4117          & 0.4396          & 0.3333          & 0.4141          \\
+ BASS w/o Augmentation & 0.3531        & 0.4047          & \textbf{0.3670} & 0.4246          \\
+ Full BASS           & \textbf{0.4120} & \textbf{0.4454} & 0.3414          & \textbf{0.4410} \\ \hline
Diffusion Policy (DP) & 0.3829          & 0.4818          & 0.3125          & 0.4526          \\
+ BASS w/o Simulation & 0.4028          & 0.5114          & 0.3569          & 0.4908          \\
+ BASS w/o Augmentation & 0.4741        & 0.5561          & 0.4200          & 0.5187          \\
+ Full BASS           & \textbf{0.5027} & \textbf{0.5707} & \textbf{0.4348} & \textbf{0.5535}
    \end{tabular}
    }\end{scalebox}
    \caption{Ablations showing the impact of each component, we show BASS with GRU and DP backbones.}
    \label{tab:ablation}
\end{table}

\section{Rollout Example}
Figure~\ref{fig:rollout_example_task_2} shows an example rollout on \updatetext{Challenge} 2. We only present one example here; additional maps and tasks can be found in the supplementary video.
\begin{figure*}[!h]
    \centering
    \includegraphics[width=1.0\linewidth]{imgs/rollout_example_single_rotation_task_2.png}
    \vspace{-1em}
    \caption{Comparison of rollouts on \updatetext{Challenge} 2 between our method (BASS), MLP, and DP. The horizontal axis denotes frames as a proxy for time. Our method successfully completes the task at frame 261, whereas both MLP and DP get stuck at an intersection. This highlights the challenge of \updatetext{Challenge} 2, where handling the large fixed-mass circular object is particularly difficult.}
    \label{fig:rollout_example_task_2}
\end{figure*}

\section{MAPPO Training Setting}
\label{appx:mappo_setting}

To train MAPPO, we integrate the Moving Out environment into the BenchMARL~\citep{bettini2024benchmarl} multi-agent RL library. Our approach to MAPPO training was designed to align with the objectives of \updatetext{Challenge} 1 and \updatetext{Challenge} 2, which were initially conceptualized with dataset-driven methods in mind. We adapted the conditions for MAPPO as follows:

For \updatetext{Challenge} 1, which originally involved training on data collected from some human players and testing on data from unseen human players, we interpret this as a zero-shot coordination challenge for MAPPO. This setup evaluates their ability to develop coordination strategies from scratch in the absence of direct human examples.

For \updatetext{Challenge} 2, the initial idea was to train on maps with diverse physical characteristics and then evaluate generalization to environments with unseen physical features. To mirror this for MAPPO, the agents are trained on maps where various physical properties (object masses, shapes, and sizes) are randomized, similar to the randomization process used during data collection for behavior cloning. Following this training phase, MAPPO's performance is then evaluated on maps with fixed physical characteristics that were not encountered during training.

\subsection{Hyperparameters}

\begin{longtable}{>{\raggedright\arraybackslash}p{0.6\textwidth} >{\centering\arraybackslash}p{0.3\textwidth}}
\caption{Summary of Parameters for MAPPO}\\
\toprule
\textbf{Parameter Name} & \textbf{Value} \\
\midrule
\endfirsthead

\multicolumn{2}{c}%
{{\bfseries \tablename\ \thetable{} -- Continued}} \\
\toprule
\textbf{Parameter Name} & \textbf{Value} \\
\midrule
\endhead

\midrule
\multicolumn{2}{r}{{Continued on next page}} \\
\endfoot

\bottomrule
\label{mappo_parameters} % Moved label to standard position for longtable
\endlastfoot

% General Training Parameters
Share Policy Parameters & True \\
Share Policy Critic & True \\

% Reinforcement Learning Algorithm Parameters
Gamma ($\gamma$)& 0.99 \\
Learning Rate & 0.00005 \\
Adam Epsilon & 0.000001 \\
Clip Gradient Norm & True \\
Clip Gradient Value & 5 \\

% Target Network Update Parameters
Soft Target Update & True \\
Polyak Tau ($\tau$) & 0.005 \\
Hard Target Update Frequency & 5 \\

% Exploration Strategy and PPO Specific Parameters
Initial Exploration Epsilon & 0.8 \\
Final Exploration Epsilon & 0.01 \\
Clip Epsilon & 0.2 \\
Critic Coefficient & 1.0 \\
Critic Loss Type & \texttt{l2} \\ % Kept l2 in texttt as it's a specific type
Entropy Coefficient & 0 \\
Lambda ($\lambda$) for GAE & 0.9 \\

% Episode and Termination Conditions
Max Cycles Per Episode & 1000 \\
Max Frames & 30,000,000 \\

% On-Policy Algorithm Specific Parameters
On-Policy Collected Frames Per Batch & 6000 \\
On-Policy Environments Per Worker & 10 \\
On-Policy Minibatch Iterations & 45 \\
On-Policy Minibatch Size & 400 \\
\midrule % Replaced \hline with \midrule for consistency with booktabs

% Model Architecture
Model Type & MLP \\
Linear Layer Sizes & [256, 256] \\
Activation Function & \texttt{torch.nn.Tanh} \\ % Kept in texttt as it's a specific function

\end{longtable}

For coordination maps, due to the greater distance from the initial explorer positions to the target items and the presence of more walls, we increased \texttt{max\_cycles\_per\_episode} from 1000 to 3000.
Concurrently, we adjusted \texttt{entropy\_coef} to 0.00065 and \texttt{gamma} to 0.92 for these maps.

\section{Reward Setting}

\subsection{Dense Reward Setting}

The dense reward is based on the change in distance $\Delta d = d_{\text{prev}} - d_{\text{curr}}$, scaled by a factor $\gamma = 20$, where $d_{\text{prev}}$ and $d_{\text{curr}}$ denote the agent’s distance to the current target at the previous and current timestep, respectively. When the agent is not holding an object, the target is either the nearest unheld item or a middle/large item currently being moved by another agent that requires assistance. When the agent is holding an object, the target becomes the goal region. At each timestep, the agent receives a reward of $\Delta d \times \gamma$.
See Tab.~\ref{tab:dense_rewards} for more details.

Additionally, there are special rewards tailored for specific maps. In Map 11 (Four Corners), for instance, two agents need to hold the two short sides of a rectangular item to more easily pass through a path successfully. Therefore, to encourage this, the reward calculation for the agents' distance to this item has been modified: instead of being based on the distance to the item's center point, it is now calculated based on the distance to its two short sides. This change is designed to encourage the agents to grasp the rectangle by its short ends.

\begin{table}[h!]
\centering
\caption{Dense Reward Settings}
\label{tab:dense_rewards}
\begin{tabular}{>{\raggedright\arraybackslash}p{0.3\textwidth} >{\raggedright\arraybackslash}p{0.4\textwidth} >{\centering\arraybackslash}p{0.2\textwidth}}
\toprule
\textbf{Primary State / Event} & \textbf{Specific Condition} & \textbf{Reward Value} \\
\midrule

% Distance-based Rewards
\multicolumn{3}{l}{\textit{A. Distance-based Rewards} } \\
\midrule
Agent not holding an item & Agent moves closer to the nearest available item & $\Delta d \times \gamma$ \\
                          & Agent moves closer to a middle or large item currently held by another agent & $\Delta d \times \gamma$ \\
\midrule
Agent holding an item     & Agent moves closer to the nearest goal region & $\Delta d \times \gamma$ \\
\midrule

% Event-based Rewards for Holding
\multicolumn{3}{l}{\textit{B. Event-based Rewards: Agent Holds an Item}} \\
\midrule
Agent successfully holds an item & Default reward for picking up & $+0.5$ \\
                                 & \textit{Exception:} If another agent is already holding other middle or large item at this time & $-0.5$ (total for this hold event) \\
                                 & \textit{Exception:} If the item picked up was already located within a goal region & $-0.5$ (total for this hold event) \\
\midrule

% Event-based Rewards for Unholding
\multicolumn{3}{l}{\textit{C. Event-based Rewards: Agent Unholds an Item}} \\
\midrule
Agent successfully unholds an item & Item is released inside a goal region & $+0.5$ \\
                                   & Item is released \textit{not} inside a goal region & $-0.5$ \\
                                   \\
                                   & \textit{Exception:} If another agent needs help, holding a large or middle item outside the goal region, at the moment of unholding. & $+0.5$ (additive) \\
\midrule

% Time-based Rewards
\multicolumn{3}{l}{\textit{D. Time-based Reward (Step Cost)}} \\
\midrule
Each timestep             & Agent exists in the environment & $-0.01$ \\
\bottomrule
\end{tabular}
\end{table}

\subsection{Does MAPPO work in Moving Out with sparse reward setting?}

The primary challenge in Moving Out lies in its significantly larger and more complex state space. Within such an expansive environment, agents who take random exploration struggle to successfully complete the multi-step tasks required to reach goal states and thus rarely receive the sparse or event-based rewards crucial for learning. Consequently, sparse reward formulations currently appear insufficient for effective policy learning in Moving Out.

MAPPO algorithms employing sparse or event-based rewards have achieved notable success in environments such as Overcooked-AI. This success can be largely attributed to the characteristics of Overcooked-AI, specifically its discrete action-state space and relatively compact overall state space. These features allow agents to encounter rewarding events with sufficient frequency through exploration, even when rewards are not dense, facilitating effective policy learning.

In Overcooked, the state-action space is small and discrete, with only tens of possible states and six possible actions, effectively rendering it a tabular setting. In contrast, our environment features continuous state and action spaces, states include precise map coordinates, and actions involve continuous control over speed and direction. Although RL is relatively easy for small discrete space, extending methods to handle continuous space is non-trivial.

Moreover, the tasks in Overcooked are relatively simple: agents fetch onions from a fixed area and deliver them using plates. Onions and plates are homogeneous, unlimited, and confined to designated regions. Once picked up, items can only be placed in predefined locations for handoff, simplifying coordination between agents.

By comparison, our tasks are significantly more complex with additional physical constraints. First, the items in our environment are heterogeneous, which are randomized in shape, size, and initial position. Thus, agents must learn to generalize over combinations of all possible scenarios. Second, unlike Overcooked, where items can only be placed in fixed zones, our agents can place items anywhere on the map. This greatly increases the difficulty of learning how to transfer items to target locations or hand them off between agents, especially in a continuous space. Additionally, our framework requires agents to engage in a wider range of collaborative behaviors beyond simple item passing—for instance, jointly moving large objects or coordinating to rotate items in tight spaces like wall corners. This diversity of collaboration types introduces further complexity.

\section{Comparative analysis of the Behaviors of BC and RL agents}
\label{appx:comparsion_BC_RL}

The fundamental difference between Behavior Cloning (BC) and MAPPO lies in their learning mechanisms and resulting agent behaviors. BC methods are inherently data-driven, leading to policies whose actions and overall effectiveness closely mirror the human behaviors captured in the training dataset. In contrast, MAPPO, as a reinforcement learning (RL) approach, develops behaviors that are strongly guided by the specific design of its dense reward function.

This distinction is evident in specific scenarios. For instance, on Map 6 (Distance Priority), both agents have their closest middle-sized items. However, human demonstration data frequently shows a strategy of first securing two smaller items before returning to move a middle-sized item together. A MAPPO agent, guided by a dense reward that incentivizes moving the nearest object, will typically prioritize the closer middle-sized item. If two such items are equidistant to respective agents (e.g., a pink agent targeting a yellow star and a blue agent targeting a blue circle), the initial actions will be independent. The coordination emerges when one agent successfully grasps a middle-sized item; the reward structure then incentivizes the other agent to assist with that specific item. Thus, the RL behavior can appear as a race to secure a primary middle-sized object, with the "loser" then being redirected by rewards to help the "winner."
BC models on Map 6 (Distance Priority), however, reflect the diversity of the human dataset. This dataset contains instances of both "small-items-first" and "middle-item-first" strategies. Consequently, a BC agent might exhibit behaviors where one agent targets a middle-sized item while the other simultaneously attempts to move a small item, reflecting a momentary misalignment as different agents emulate distinct strategies observed in the human data.

Map 11 (Four Corners) further illustrates these differences. Here, two agents might each have two items at an equal distance, making multiple initial moves potentially optimal. In our MAPPO training, agents often exhibit initial movements that appear somewhat exploratory or randomized until one agent commits to and grasps a large item. At this point, the dense reward system effectively directs the other agent to provide assistance.
Conversely, BC models on Map 11 (Four Corners) tend to display more decisive and rapidly aligned behavior from the start. Observations of the human dataset for this map revealed a common leader-follower dynamic, where one player (e.g., the blue agent) consistently follows the lead of the other (e.g., the pink agent). If the pink agent, for example, decisively moves towards an upper pink square, the blue agent often follows suit immediately to assist. As a result, BC models rarely exhibit prolonged periods of uncoordinated or hesitant movement before aligning on a common goal.

In summary, BC methods excel at reproducing observed human behaviors, including their specific strategies and inherent diversity. RL approaches like MAPPO, while capable of discovering effective strategies, are highly sensitive to the nuances of reward function design. Even slight modifications to the reward signals can lead to significant and sometimes qualitatively different emergent behaviors in the trained agents.

\section{Comparison with zero-shot coordination (ZSC)}
\updatetwotext{
We compare DP/BASS with representative multi-agent RL and zero-shot coordination baselines, including MAPPO and CoMeDi~\cite{sarkar2023diverse}, to further evaluate generalization to unseen human proxies. The comparison is conducted on two representative layouts, Map 1 and Map 12. The task completion rates (TCR) are reported in Table~\ref{tab:zsc_comparison}.}

\begin{table}[h]
\centering
\caption{Comparison with ZSC baselines on seen and unseen human proxies.}
\label{tab:zsc_comparison}
\begin{tabular}{l c c c c}
\toprule
Method & Map 1 Seen & Map 1 Unseen & Map 12 Seen & Map 12 Unseen \\
\midrule
MAPPO & 0.5842 & 0.3287 & 0.4076 & 0.1750 \\
CoMeDi~\cite{sarkar2023diverse} & 0.6278 & 0.3768 & 0.4734 & 0.2570 \\
DP/BASS & $\mathbf{0.7868}$ & $\mathbf{0.7080}$ & $\mathbf{0.5993}$ & $\mathbf{0.3855}$ \\
\bottomrule
\end{tabular}
\end{table}

\updatetwotext{The results show that ZSC methods have a clear performance drop when paired with unseen human proxies. For example, CoMeDi decreases from 0.6278 to 0.3768 on Map 1, and from 0.4734 to 0.2570 on Map 12. In contrast, DP/BASS achieves higher TCR on both seen and unseen human proxies, indicating stronger robustness to unseen collaborative behaviors.}

\updatetwotext{This performance gap suggests a limitation of directly applying ZSC methods to continuous physical collaboration environments such as Moving Out. In Moving Out, RL-based ZSC methods require dense rewards to converge reliably. However, dense rewards can also encourage trajectory-level convergence across different behavioral conventions. As a result, although high-level choices, such as selecting the left or right path, may remain diverse, the low-level continuous motion patterns become homogeneous. This reduces the agent's ability to adapt to unseen human physical behaviors. We also evaluated event-based and sparse reward variants, but they did not converge reliably in this environment. Overall, these results suggest that data-driven behavior modeling, as used in BASS, is important for robust collaboration in continuous physical environments.}

% \textbf{ZSC performance drops on unseen behaviors.}
% \updatetwotext{While ZSC methods such as CoMeDi can perform well in self-play, their performance drops substantially when paired with unseen human proxies. For example, CoMeDi drops from 0.6278 to 0.3768 on Map 1. In contrast, BASS remains significantly more robust to unseen behaviors.}

% \textbf{Why ZSC struggles in Moving Out.}
% \updatetwotext{Our analysis suggests a fundamental bottleneck for applying ZSC methods to continuous physical collaboration environments such as Moving Out. RL methods require dense rewards to converge reliably in this environment. However, dense rewards can also drive trajectory-level convergence across different conventions. As a result, although high-level options, such as choosing the left or right path, may remain diverse, low-level continuous motion patterns become homogeneous. This limits the agent's ability to align with unseen human physical actions. We also tested event-based and sparse rewards, but they failed to converge reliably.}

% In summary, these results show that data-driven behavior modeling, as used in BASS, is important for continuous physical collaboration.

\section{Additional Validation of Behavior Augmentation}
\label{appx:validation}

\textbf{Behavior mismatch in sub-trajectory recombination.}
Our recombination strategy is explicitly designed to avoid the type of inconsistency that was described. Specifically, we only perform sub-trajectory swapping when the fixed agent (e.g., agent A) has the same start and end poses across two trajectories. This ensures that agent A is pursuing the same local goal in both cases, regardless of the specific behavior of the partner.

For example, suppose in trajectory $\tau_{1}$, agent A performs action sequence $a_{1}$ while agent B performs $b_{1}$, and in $\tau_{2}$, A performs $a_{2}$ while B performs $b_{2}$. If $a_{1}$ and $a_{2}$ share the same start and end states, we can create two new combinations: $(a_{1}, b_{2})$ and $(a_{2}, b_{1})$. These are valid because both $b_{1}$ and $b_{2}$ were originally compatible with different variants of A’s strategy toward the same goal. As such, swapping B's behavior does not interfere with A’s intent. This preserves behavioral diversity while ensuring trajectory-level coherence.

\medskip
\noindent
\textbf{Additional validation.}
To ensure consistency, we identify sub-trajectories with matching start and end poses, so that the recombined agent behaviors maintain the same intention and goal. We have validated this approach with over 99\% success rate in producing physically valid trajectories. In addition, we also confirm that the diversity of behavior increases after augmentation (Entropy improves from 0.88 to 0.95).

\section{Details of Evaluation Metrics}
\label{appx:evaluation_metrics}

To assess human-AI collaboration in \textit{Moving Out}, we design metrics that go beyond final task success to capture the \emph{quality of physical collaboration}. 
While prior works such as Overcooked-AI mainly rely on task completion, this is insufficient in physically grounded settings, where interactions involve continuous control, object dynamics, and force alignment. 
We therefore complement \textbf{Task Completion Rate (TCR)} with three additional metrics---\textbf{Normalized Final Distance (NFD)}, \textbf{Waiting Time (WT)}, and \textbf{Action Consistency (AC)}---each targeting a different aspect of collaboration under physical constraints.

\paragraph{Task Completion Rate (TCR).}  
TCR measures the proportion of objects successfully delivered to goal regions, weighted by size:
\[
TCR = \frac{\sum w_i \mathbb{I}(o_i \text{ delivered})}{\sum w_i},
\]
where \(w_i = 1\) (small) or \(2\) (middle/large). Range: [0,1].  
TCR captures the final outcome of collaboration, but by itself cannot distinguish between failed attempts with meaningful progress and those with no progress.

\paragraph{Normalized Final Distance (NFD).}  
NFD quantifies partial progress by measuring the reduction in object-goal distance:
\[
NFD = 1 - \frac{\sum_{i=1}^{N} d_i^{\text{final}}}{\sum_{i=1}^{N} d_i^{\text{initial}}},
\]
where \(d_i^{\text{initial}}\) and \(d_i^{\text{final}}\) are the object's initial and final distances to the target.  
This is critical in physical environments where objects may get stuck due to collisions or narrow passages. 
A case with high NFD but low TCR indicates that agents made progress but failed to overcome physical constraints.

\paragraph{Waiting Time (WT).}  
WT captures how agents coordinate when joint effort is required:
\[
WT = \sum_{t \in \mathcal{W}} (t_{\text{end}}^t - t_{\text{start}}^t),
\]
where \(\mathcal{W}\) is the set of intervals when an agent holds a middle/large object but must wait for help.  
High WT may reflect either poor recognition of the need for help or inefficiency in navigating physical obstacles. 
Thus, it measures awareness and responsiveness in physically grounded collaboration.

\begin{figure*}[!h]
    \centering
    \includegraphics[width=0.6\linewidth]{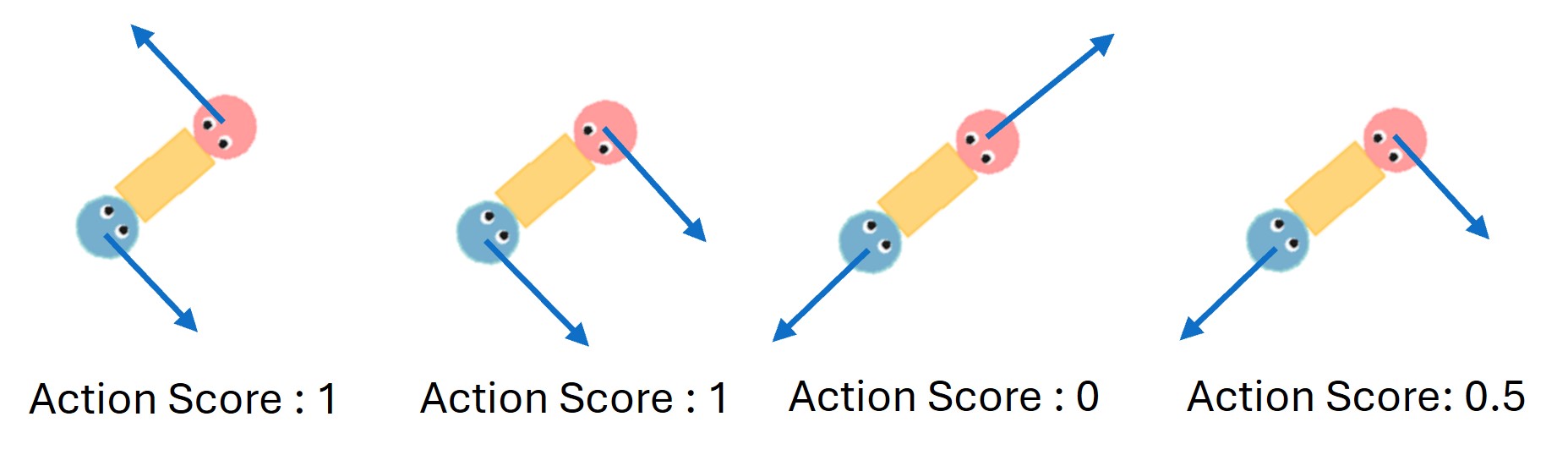}
    \vspace{-1em}
    \caption{Example of action consistency (AC) calculation. Effective collaborative work receives high scores, while opposing forces canceling each other lead to low scores.}
    \label{fig:ac_calcualtion_example}
\end{figure*}

\paragraph{Action Consistency (AC).}  
As illustrated in Fig.~\ref{fig:ac_calcualtion_example}, AC measures how well two agents align their applied forces during joint manipulation:
\[
AC = \frac{1}{T} \sum_{t=0}^{T-1} \frac{\| (\vec{f}_1^t + \vec{f}_2^t) \cdot \vec{d}_t \|}{\| \vec{f}_1^t \| + \| \vec{f}_2^t \|},
\]
where \(\vec{f}_1^t, \vec{f}_2^t\) are the forces applied at time \(t\), \(\vec{d}_t\) is the unit vector connecting agent positions, and \(T\) is the number of timesteps.  
This metric captures coordination quality: agents are most effective when their forces are aligned, and receive low scores when their efforts cancel out (e.g., one pushing forward, the other pulling back).

Together, these four metrics provide a comprehensive evaluation of human-AI collaboration in physically grounded tasks:  
TCR reflects task success, NFD measures partial progress under constraints, WT captures coordination in joint effort, and AC quantifies the efficiency of physical interaction. 
This combination moves beyond symbolic settings and offers a richer view of how collaboration unfolds in continuous, low-level environments.

\section{Implementation Details}
\label{appx:implementation_details}
\subsection{Environment Details}

\subsubsection{Observation Encoding}

\paragraph{State Observation}

Our observation encoding is ego-centric and represents all information as a one-dimensional vector. The encoded information includes:

\begin{itemize}
    \item Self: Position and angle, with angles $\theta$ represented using \([\cos\theta, \sin\theta]\). A boolean value indicates whether the agent is holding an item (True/False).  
    \item Partner: Position, angle, and whether it is holding.  
    \item Items: Each item is encoded with position, angle, size, category, and shape. Category and shape use a one-hot encoding.  
\end{itemize}

When training on a single map, the walls and goal region remain unchanged, so we do not encode them. However, when training across different maps, we include their encoding:

\begin{itemize}
    \item Walls: Represented by the \((x, y)\) coordinates of the top-left and bottom-right corners.  
    \item Goal Region: Represented the save as walls. The top-left and bottom-right corners.  
\end{itemize}

% \paragraph{Image Observation}

% When using an image as the observation, the agent appears blue in its own view, while the partner appears pink. This allows both agents to have ego-centric observations. The observation image is a \(256 \times 256\) RGB image.

\subsubsection{Action Encoding}
The agent's action space has four values:
\begin{itemize}
    \item The movement distance (forward or backward).
    \item The target angle (encoded using $\cos$ and $\sin$).
    \item The grasping action: $1$ means grasp or release, $0$ means no change.
\end{itemize}

% \subsubsection{Evaluation Setting}

\subsection{Baseline Details}

\begin{itemize}
    \item \textbf{Diffusion Policy}: We follow the original implementation by \citep{chi2023diffusion} for the model architecture, which employs a 1D U-Net to generate action sequences. The observation, prediction, and executable horizons are set to 2, 8, and 4, respectively. Training is performed using the Adam optimizer with 1k epochs, 1024 batch size, and 0.001 learning rate. The diffusion steps are 36. The grasp action is encoded by one-hot encoding.
    \item \textbf{MLP} The MLP model consists of 3 fully connected layers with Tanh activation and hidden\_dim 2048. It concatenates one past state and one current state as input and predicts actions for the next 8 steps. Training is performed using the Adam optimizer with 1k epochs, 1024 batch size, and 0.001 learning rate. It optimizes a combination of mean squared error (MSE) loss for movement outputs and cross-entropy loss for grasp action predictions.
    \item \textbf{GRU} uses a GRU layer followed by 3 fully connected layers with Tanh activation and hidden\_dim 2048. It takes one past state and the current state as input and predicts actions for the next 4 steps. The model processes sequential data and learns action patterns based on previous movements. Training is performed using the Adam optimizer with 1k epochs, 1024 batch size, and 0.001 learning rate. It optimizes a combination of mean squared error (MSE) loss for movement outputs and cross-entropy loss for grasp action predictions.
    
\end{itemize}

\subsection{BASS Details}
\label{appx:bass_details}
\begin{itemize}
    \item \textbf{Dynamics Model} The Autoencoder consists of an encoder and a decoder, both made of two linear layers. They use ReLU as the activation function, and each layer has 128 units. The latent space has 32 dimensions. The dynamics Model is a two-layer MLP (Multi-Layer Perceptron). Each hidden layer has 128 units. During training, the two autoencoders and the dynamic model are trained together. Additionally, we also explored fine-tuning the second AE from the first.  Our ablation on selected Maps 2, 6, \& 9 shows the following average NFDs: 1) Joint training: 0.55, 2) Fine-tuning the second AE from the first AE: 0.50, 3) Training two AEs separately: 0.48.

    \item \textbf{Partner Action Predictor} The Partner Action Predictor can be designed based on the application. In some cases, it can be the same as the action policy, but with a small change: it swaps the agent's state with the partner's state. This allows the model to predict the partner’s action from their perspective.
\end{itemize}

\paragraph{Behavior Augmentation and Recombination Sub-Trajectories}

In behavior augmentation, we add noise with a mean of $0$ and a standard deviation of $0.002$.  
In recombination sub-trajectories, since two points in a continuous space are almost never the same, we set a tolerance value.  
We discretize the environment into a $48 \times 48$ grid. If the robot's start and end points are in the same grid cell, we treat them as the same point.

\paragraph{Normalized Final Distance Calculation}

Many maps have walls, so we cannot use Euclidean distance.  
To improve efficiency, we discretize the environment into a $48 \times 48$ grid.  
We use the BFS algorithm to compute the distance from the item to the Goal Region.

\paragraph{Can BASS be used as a standalone Method beyond Moving Out?}

BASS is a standalone method composed of two components. Together, they make BASS applicable across various behavior cloning methods outside of Moving Out, as discussed.
We tested BASS on a widely used human-AI collaboration environment, Overcooked AI, specifically, the "Cramped Room" map. The results showed that DP+BASS improved the score by ~15\% compared to DP alone. This demonstrates that BASS is not limited to Moving Out.

\section{Scalability to more complex environments.}

\textbf{Higher-dimensional observations and more objects.}
BASS can handle more complex state spaces and more objects by replacing the MLP encoder with a visual encoder, such as a CNN. We evaluated Map 6 using raw image inputs and our image-based progress estimator, as described in Appendix~\ref{appx:progress_TCL_esti}. The results are shown in Table~\ref{tab:image_bass}.

\begin{table}[h]
\centering
\caption{BASS with state-based and image-based observations on Map 6.}
\label{tab:image_bass}
\begin{tabular}{l c c}
\toprule
Method & Challenge 1 TCR ($\uparrow$) & Challenge 2 TCR ($\uparrow$) \\
\midrule
DP (State) & 0.4106 & 0.2935 \\
DP/BASS (State) & $\mathbf{0.5812}$ & $\mathbf{0.3907}$ \\
DP (Image) & 0.3069 & 0.2225 \\
DP/BASS (Image) & $\mathbf{0.3417}$ & $\mathbf{0.2516}$ \\
\bottomrule
\end{tabular}
\end{table}

Although raw image inputs naturally reduce the overall performance, BASS consistently outperforms the DP baseline. This shows that BASS can effectively scale to high-dimensional visual observation spaces.

\textbf{More agents.}
Most modules in BASS, including noise perturbation, learned dynamics, and action selection, can scale to more agents by adding a predictor for each new teammate. However, sub-trajectory recombination becomes a bottleneck. This module requires the start and end positions of all agents to match simultaneously. As the number of agents increases, the probability of finding such perfect matches drops sharply, making direct recombination difficult in many-agent settings.

\section{Study on Action Sampling Times}
\label{appx:bass_sampling}

We conducted an ablation on the number and strategy of action samples used in BASS. 
In our implementation, each candidate action is generated by independently sampling from the policy and partner predictor up to four times, producing four simulation rollouts. 
This setting provides a good balance between performance and efficiency, supporting real-time human evaluation at 10Hz.

To compare alternatives, we tested three strategies:  
1) Independent 4× sampling (current setting);  
2) 2×2 combination (two samples from each agent, combined into four rollouts);  
3) 4×4 combination (four samples from each, combined into sixteen rollouts).  

\begin{table}[h]
\centering
\resizebox{0.8\textwidth}{!}{
\begin{tabular}{lcccccccc}
\toprule
 & \multicolumn{4}{c}{\updatetext{Challenge} 1} & \multicolumn{4}{c}{\updatetext{Challenge} 2} \\
\cmidrule(lr){2-5} \cmidrule(lr){6-9}
 & TCR$\uparrow$ & NFD$\uparrow$ & WT$\downarrow$ & AC$\uparrow$ 
 & TCR$\uparrow$ & NFD$\uparrow$ & WT$\downarrow$ & AC$\uparrow$ \\
\midrule
Sample 4$\times$ independently  & 0.5027 & 0.5707 & 0.3448 & 0.8615 & 0.4348 & 0.5535 & 0.3096 & 0.8474 \\
2$\times$2 combination          & 0.4522 & 0.5261 & 0.3485 & 0.8462 & 0.4158 & 0.5137 & 0.3101 & 0.8460 \\
4$\times$4 combination          & \textbf{0.5161} & \textbf{0.5847} & \textbf{0.3390} & \textbf{0.8727} 
                                & \textbf{0.4396} & \textbf{0.5638} & \textbf{0.3087} & \textbf{0.8559} \\
\bottomrule
\end{tabular}
}
\caption{Results of sampling and combination strategies.}
\label{tab:sampling_comparison}
\end{table}

Results are summarized in Table~\ref{tab:sampling_comparison}. The 2×2 strategy, despite using the same number of simulations as the independent setting, consistently underperforms. Independent sampling has higher chance to capture critical joint transitions, e.g., resolving a stuck state. The 4×4 combination achieves the best accuracy, but requires 16 rollouts and increases inference time from 69\,ms to 210\,ms, which disrupts real-time human evaluation at 10Hz. 

We therefore adopt the independent 4× sampling scheme in BASS, as it balances accuracy with the real-time feasibility required for human-in-the-loop collaboration.

\section{BASS Analysis}
\subsection{BASS Module Analysis}
\label{app:bass_module_analysis}

\subsubsection{Next-State Prediction Accuracy}
We evaluate the accuracy of the next-state prediction module by computing the L2 distance between the predicted state and the ground-truth state from the oracle simulator. Two baselines are included: a GRU-based predictor and a random state generator.

\begin{table}[h]
\centering
\begin{tabular}{lcc}
\toprule
L2 Norm (↓) & \updatetext{Challenge} 1 & \updatetext{Challenge} 2 \\
\midrule
BASS          & \textbf{0.0010} & \textbf{0.0028} \\
GRU           & 0.0196          & 0.0331          \\
Random States & 2.7576          & 3.2594          \\
\bottomrule
\end{tabular}
\caption{Next-state prediction accuracy across tasks. Lower is better.}
\label{tab:next_state_pred}
\end{table}

These results show that BASS more accurately captures physical dynamics compared to both the GRU baseline and random guessing, supporting the effectiveness of the learned dynamics model in simulation and action selection.

\subsubsection{Modeling Diverse Human Behaviors}
To handle diverse human behaviors, our approach models partner actions as a conditional distribution learned from demonstrations. The latent dynamics model captures this diversity by representing multiple likely behaviors under the same state, instead of committing to a single mode.

We use a Diffusion Policy, which effectively models multimodal action distributions by sampling from different noise inputs~\citep{li2024learning,chi2023diffusion}. This enables the model to generate different possible partner responses, providing probabilistic reasoning that aligns with human intuition.

\subsubsection{Partner Action Prediction Accuracy}
Since the action space is continuous, prediction accuracy is evaluated using the relative error between predicted and ground-truth actions. With a 10\% error tolerance, the predictor achieves an accuracy of 71.45\%; relaxing the tolerance to 20\% increases accuracy to 90.24\%. These results indicate that the partner action predictor provides sufficiently accurate estimates to support effective next-state prediction and action selection within our framework.

\subsubsection{Effect of Random Partner Actions}
To assess the importance of accurate partner action prediction, we conduct an ablation where the partner's actions are randomly sampled during the simulation step. Since our action selection process considers four candidates, random guessing introduces uncertainty and degrades the ability to make correct selections between predicted future states.

\begin{table}[h]
\centering
\begin{tabular}{lcccc}
\toprule
 & \multicolumn{2}{c}{TCR (↑)} & \multicolumn{2}{c}{Action Selection Accuracy (↑)} \\
\cmidrule(lr){2-3} \cmidrule(lr){4-5}
 & \updatetext{Challenge} 1 & \updatetext{Challenge} 2 & \updatetext{Challenge} 1 & \updatetext{Challenge} 2 \\
\midrule
BASS (Ours)                  & 0.5027 & 0.4348 & 0.6250 & 0.4870 \\
BASS (Random partner action) & 0.3966 & 0.3650 & 0.2542 & 0.2484 \\
BASS w/ Oracle Simulator     & 0.5421 & 0.5334 & 1.0000 & 1.0000 \\
\bottomrule
\end{tabular}
\caption{Impact of replacing the learned partner model with random actions. Both task performance (TCR) and action selection accuracy drop significantly.}
\label{tab:random_partner}
\end{table}

\subsubsection{Comparison with Alternative Next-State Prediction Models}
To evaluate the effectiveness of our proposed dynamics model, we compare it with two alternatives: a GRU-based predictor and a qVAE model for next-state prediction. As shown in Table~\ref{tab:next_state_models}, our model significantly outperforms both in prediction accuracy and final task performance (NFD). The GRU baseline performs close to random guessing, indicating difficulty in learning accurate transition dynamics. The qVAE model performs slightly better, but still struggles, likely due to the large continuous state space, where discretized latent codes are insufficient to represent fine-grained physical interactions. These results highlight the importance of our autoencoder-based latent dynamics model in capturing physical transitions effectively.

\begin{table}[h]
\centering
\begin{tabular}{lcccc}
\toprule
& \multicolumn{2}{c}{NFD (↑)} & \multicolumn{2}{c}{Prediction Accuracy (↑)} \\
\cmidrule(lr){2-3} \cmidrule(lr){4-5}
& \updatetext{Challenge} 1 & \updatetext{Challenge} 2 & \updatetext{Challenge} 1 & \updatetext{Challenge} 2 \\
\midrule
BASS (Ours)            & 0.5733 & 0.5535 & 0.6250 & 0.4870 \\
BASS (GRU)             & 0.5048 & 0.4965 & 0.2487 & 0.2598 \\
BASS (qVAE)            & 0.5113 & 0.5032 & 0.3102 & 0.2722 \\
BASS w/ Oracle Simulator & 0.5875 & 0.6209 & N/A & N/A \\
\bottomrule
\end{tabular}
\caption{Comparison of next-state prediction models. Our dynamics model outperforms GRU and qVAE in both prediction accuracy and task performance.}
\label{tab:next_state_models}
\end{table}

\subsection{Comparison with Single Agent BASS}
\label{appx:comparsion_single_agent}
Our method extends augmentation, simulation, and action selection to a multi-agent collaborative setting, which introduces challenges fundamentally different from single-agent environments. In collaborative manipulation, both agents jointly affect the shared object, so a valid sub-trajectory must be compatible with the partner’s motion. By contrast, single-agent recombination only checks whether a segment starts or ends from a similar state, without ensuring that the segment represents the same behavioral goal. As formally defined in the Appendix, the single-agent version treats two segments as compatible if either \(s^i_{t_1} \approx \hat{s}^i_{t_3}\) or \(s^i_{t_2} \approx \hat{s}^i_{t_4}\). Because this criterion ignores the evolution of the partner’s motion and does not ensure that the two segments correspond to the same part of the task, the recombined trajectories can easily violate the coordinated patterns required for joint manipulation.

Our multi-agent recombination module avoids this issue by requiring that agent~\(i\) begins and ends the segment in nearly the same physical situation in both demonstrations. This ensures that agent~\(i\) is performing the same portion of the task, so the partner’s subsequence can be safely swapped while keeping agent~\(i\)’s behavior consistent throughout the segment. As shown in Table~\ref{tab:w1_recombination}, the single-agent recombination baseline increases diversity metrics but degrades collaboration performance, indicating that the generated trajectories no longer reflect valid joint behaviors. In contrast, our multi-agent recombination increases both diversity and cooperative performance, demonstrating that preserving cross-agent compatibility is essential for effective augmentation in collaborative settings.

\begin{table}[!h]
\centering
\caption{Comparison between multi-agent recombination and single-agent recombination on Moving Out \updatetext{Challenge} 1. Multi-agent recombination increases diversity while preserving coordinated behavior, whereas single-agent recombination increases diversity but harms collaboration performance.}
\label{tab:w1_recombination}
\resizebox{1.0\textwidth}{!}{
\begin{tabular}{lcccccccc}
\toprule
 & \multicolumn{4}{c}{\textbf{Diversity}} & \multicolumn{4}{c}{\textbf{Performance}} \\
\cmidrule(lr){2-5}\cmidrule(lr){6-9}
\textbf{Method} 
& DTW Mean & DTW Var & Entropy (KDE) & Coverage (RBF)
& TCR(↑) & NFD(↑) & WT(↓) & AC(↑) \\
\midrule
Multi-agent Recombination (Ours) 
& 7.526 & 6.612 & 0.892 & 0.910 
& \textbf{0.403} & \textbf{0.511} & \textbf{0.308} & \textbf{0.840} \\
Single-agent Recombination 
& \textbf{7.741} & \textbf{6.722} & \textbf{0.897} & \textbf{0.919} 
& 0.368 & 0.451 & 0.338 & 0.839 \\
No Recombination 
& 7.013 & 6.065 & 0.888 & 0.899  
& 0.383 & 0.482 & 0.345 & 0.824 \\
\bottomrule
\end{tabular}
}
\end{table}

We further examine the action simulation and selection module. If the next-state prediction model considers only one agent’s future action while ignoring the partner’s state evolution, its effectiveness is greatly diminished. As shown in Table~\ref{tab:w1_simulation}, a single-agent simulation model yields performance close to disabling simulation entirely. This highlights the importance of modeling the coupled dynamics of both agents and further confirms that the proposed approach cannot be reduced to a combination of single-agent components.

\begin{table}[!h]
\centering
\caption{Comparison of multi-agent vs. single-agent action simulation on Moving Out \updatetext{Challenge} 2. Without considering the partner’s future state, the benefit of simulation is greatly reduced.}
\label{tab:w1_simulation}
\begin{tabular}{lcccc}
\toprule
Method & TCR(↑) & NFD(↑) & WT(↓) & AC(↑) \\
\midrule
Multi-agent Action Simulation (Ours)
& \textbf{0.420} & \textbf{0.554} & \textbf{0.310} & \textbf{0.848} \\
Single-agent Action Simulation
& 0.319 & 0.458 & 0.327 & 0.833 \\
No Action Simulation
& 0.313 & 0.452 & 0.335 & 0.821 \\
\bottomrule
\end{tabular}
\end{table}

Overall, these results demonstrate that our method must explicitly account for multi-agent coordination constraints. The approach is not a direct extension of single-agent techniques but instead relies on mechanisms specifically designed to maintain cross-agent alignment. While formal theoretical guarantees are beyond the scope of this work, the empirical evidence highlights the importance of multi-agent structure, and we plan to investigate deeper theoretical characterizations in future work.

\subsection{Pose-noise augmentation without re-simulation.}
\label{appx:pose_noise_anay}
\updatetwotext{In behavior augmentation, if the pose shift is too large, the following trajectory may become physically invalid and re-simulation would be needed. To validate whether small pose noise can be added without re-simulating the full trajectory as implemented in BASS, we evaluate how much noise can be added while keeping the trajectory physically valid. }

\updatetwotext{We tested different noise levels ($\sigma \in \{0.2, 0.02, 0.002, 0.0002\}$) on Map 10 (Single Rotation). For the 80 trajectories in Challenge 1, we sampled 2,000 times by adding noise to a state and using the simulator to play the subsequent actions to check whether the trajectory remained physically valid.}

\begin{table}[h]
\centering
\caption{Ablation of different pose-noise levels on Map 10 (Single Rotation).}
\label{tab:pose_noise_ablation}
\begin{tabular}{c c c c}
\toprule
Noise ($\sigma$) & Pose Shift & Validity Rate ($\uparrow$) & TCR ($\uparrow$) \\
\midrule
0.0002 & $<0.04\%$ & $99.92\%$ & 0.5492 \\
0.002 (Current Choice) & $<0.4\%$ & $\mathbf{99.87\%}$ & $\mathbf{0.6538}$ \\
0.02 & $<4.0\%$ & $35.01\%$ & 0.0000 \\
0.2 & $<40.0\%$ & $0.02\%$ & 0.0000 \\
\bottomrule
\end{tabular}
\end{table}

\updatetwotext{Our chosen noise level ($\sigma=0.002$) has a 99\% confidence interval of roughly $[-0.005, 0.005]$, which corresponds to about a $0.4\%$ position error in the environment. As shown in Table~\ref{tab:pose_noise_ablation}, this amount of deviation preserves a $99.87\%$ physical validity rate while achieving the highest task completion rate (TCR). When the noise is too large, e.g., $\sigma=0.02$, the validity rate drops to $35.01\%$ and the task fails with $\mathrm{TCR}=0$.}

\updatetwotext{Although re-simulation guarantees a 100\% validity rate, it introduces unacceptable computational overhead. Running the physics engine to validate or re-simulate every trajectory segment for every batch during training would make training impractically slow. Since our chosen $\sigma=0.002$ already achieves a $99.87\%$ natural validity rate, heavy re-simulation is unnecessary for maintaining feasibility.}

\updatetwotext{The augmented states stay close to valid trajectories. With $\sigma=0.002$, the pose shift is less than $0.4\%$, and more than $99.8\%$ of the tested trajectories remain valid. Therefore, the noise is unlikely to cause large error accumulation. Instead, it helps the policy learn to handle small pose variations, such as minor human motion jitter, and works as a regularizer.}

\section{Evaluation Protocol for \updatetext{Challenge} 1}
\label{appx:evaluation_protocol}

\textbf{Human-based evaluation.} 
Since \updatetext{Challenge} 1 is designed to evaluate adaptation to unseen human behaviors, our primary setup requires agents to play with new human participants. 
While this is the most direct evaluation of human-AI collaboration, it raises concerns about reproducibility because new participants are required for each run. 

\textbf{Cross-group reproducible protocol.} 
To address this, we design a reproducible protocol inspired by cross-model evaluation. 
We randomly split the human demonstrations into two groups, each containing data from different participants. 
Two models are trained separately on each group and then evaluated by playing with each other. 
This setup emulates collaboration with unseen partner behavior while remaining fully reproducible. 

\textbf{Results.} 
Table~\ref{tab:cross_group_eval} compares the performance of Diffusion Policy (DP) and our method (BASS) when paired with models trained on the same group vs.\ a different group. 
We also report the percentage drop (or increase) in performance when moving from same-group to cross-group evaluation. 
Results show that models perform better when paired with a model trained on the same group (more aligned behavior), but BASS consistently outperforms DP when paired with unseen human behaviors, demonstrating stronger generalization.

\begin{table}[h]
    \centering
    
    \vspace{0.5em}
    \resizebox{1.0\textwidth}{!}{
    \begin{tabular}{l|l|c|c|c|c}
        \toprule
        Setting & Method & TCR (↑) & NFD (↑) & WT (↓) & AC (↑) \\
        \midrule
        Same-Group & DP      & 0.3233 & 0.5367 & 0.3789 & 0.8163 \\
                   & DP/BASS & 0.3503 & 0.5724 & 0.3598 & 0.8337 \\
        \midrule
        Cross-Group & DP      & 0.2563 (-20.72\%) & 0.4589 (-14.50\%) & 0.4249 (+12.15\%) & 0.7854 (-3.78\%) \\
                    & DP/BASS & 0.3010 (-14.07\%) & 0.5197 (-9.22\%)  & 0.3899 (+8.37\%)  & 0.8099 (-2.86\%) \\
        \bottomrule
    \end{tabular}
    }
    \caption{Cross-group evaluation protocol for \updatetext{Challenge} 1. Performance drops (\%) are measured relative to same-group evaluation.}
    \label{tab:cross_group_eval}
\end{table}

\section{Why choose Normalized Final Distance in BASS?}
\label{appx:choose_NFD}
We chose Normalized Final Distance (NFD) because it directly reflects task progress in physically grounded collaboration. When two agents move an object together, actions that successfully reduce the distance between objects and the goal region indicate effective cooperation, even when navigating around obstacles like walls. Thus, maximizing NFD considers both physical feasibility and cooperation efficiency.

We also experimented with a multi-objective scoring using both NFD and Action Consistency (AC), to encourage not only progress but also force alignment. The trade-off is shown below:

\begin{table}[h]
\centering
\begin{tabular}{lcc}
\toprule
\updatetext{Challenge} 1 & NFD ($\uparrow$) & AC ($\uparrow$) \\
\midrule
BASS with NFD     & 0.5733 & 0.8615 \\
BASS with NFD+AC  & 0.5683 & 0.9127 \\
\bottomrule
\end{tabular}
\caption{Comparison of using NFD vs. NFD+AC as objectives.}
\label{tab:nfd_vs_ac}
\end{table}

Although this combined objective improved AC, we found that NFD dropped slightly. We chose to prioritize NFD in the paper for its ability to capture physical task progress.

\section{Behavior Augmentation Details}
\label{appx:behavior_aug_details}
Our augmentation strategy involves two techniques:

\textbf{Generating New States by Perturbing the Partner's Pose}
For a given trajectory, we generate new states by introducing noise to the partner's pose while keeping all other state variables unchanged.
This perturbation creates additional observation variations in training data, allowing the agent to experience a broader range of possible partner behaviors.
Since human actions naturally vary, this approach helps improve the agent's robustness to small deviations in the partner's movements while maintaining its own task objectives.
This perturbation is expressed as $
\tilde{p}_{\text{partner}} = p_{\text{partner}} + \epsilon, \epsilon \sim \mathcal{N}(0, \sigma^2) $
where \( p_{\text{partner}} \) is the original partner's pose, \( \epsilon \) is Gaussian noise with mean \( 0 \) and variance \( \sigma^2 \), and \( \tilde{p}_{\text{partner}} \) is the perturbed pose used to generate new state variations.

\textbf{Recombination of Sub-Trajectories}
Each global state $s_t$ can be decomposed into
\(
s_t \;=\; \bigl(s^i_t,\; s^j_t,\; s^e_t\bigr),
\)
where $s^i_t$ and  $s^j_t$ are the individual states of agent $i$ and $j$, and $s^e_t$ captures the remaining environment-specific information.
Given a trajectory $\tau = \left\{(s_t, a_t)\right\}_{t=1:T}$, we extract three sequences: $\tau^i = \left\{(s_t^i, a_t^i)\right\}_{t=1:T}$ is the state-action sequence of agent $i$; similarly $\tau^j$ is the state-action sequence of agent $j$ and $\tau_e$ is the sequence of environment information.
We have $\tau = \tau^i\cup \tau^j \cup \tau^e$.
Moreover, let $\tau_t^i = (s_t^i, a_t^i)$ be the $t$-th state-action pair of agent $i$, and define $\tau_{t_1:t_2}^i = (s_{t_1}^i, a_{t_1}^i, \cdots, s_{t_2}^i, a_{t_2}^i)$ as the continuous sub-trajectory of $\tau^i$ from $t_1$ to $t_2$.
We can define $i$'s trajectory composed of sub-trajectories $\tau^i = \tau^i_{1:t_1 - 1} \cup \tau^i_{t_1:t_2}  \cup \tau^i_{t_2+1:T}$; and similarly for $j$.

Given $\tau$ and two time step $t_1, t_2$, we can search for another trajectory $\hat{\tau}$ in the dataset such that $\hat{\tau}_{t_1}^i = \tau_{t_1}^i$ and $\hat{\tau}_{t_2}^i = \tau_{t_2}^i$.
We can then construct two new trajectories by swapping agent $j$'s subsequences between $t_1$ and $t_2$:
\begin{equation*}
    \tau^i \cup \left(\tau^j_{1:t_1 - 1} \cup \hat{\tau}^j_{t_1:t_2}  \cup \tau^j_{t_2+1:T}\right) \cup \tau^e
    \;\;\;\text{and}\;\;\; 
    \hat{\tau}^i \cup \left(\hat{\tau}^j_{1:t_1 - 1} \cup \tau^j_{t_1:t_2}  \cup \hat{\tau}^j_{t_2+1:T}\right) \cup \hat{\tau}^e
\end{equation*}
%\vspace{-2em}

By aligning the start and end of agent~$i$'s sub-trajectory, the generated trajectories maintain temporal consistency for agent~$i$ while introducing a different  sequence.
This approach enriches the training set with new, valid trajectories where agent~$i$’s behavior is fixed and the partner’s varies.

\subsection{Visualization Example of Recombination}
\label{appx:visal_recomb}
\begin{figure*}[!h]
    \centering
    \includegraphics[width=0.8\linewidth]{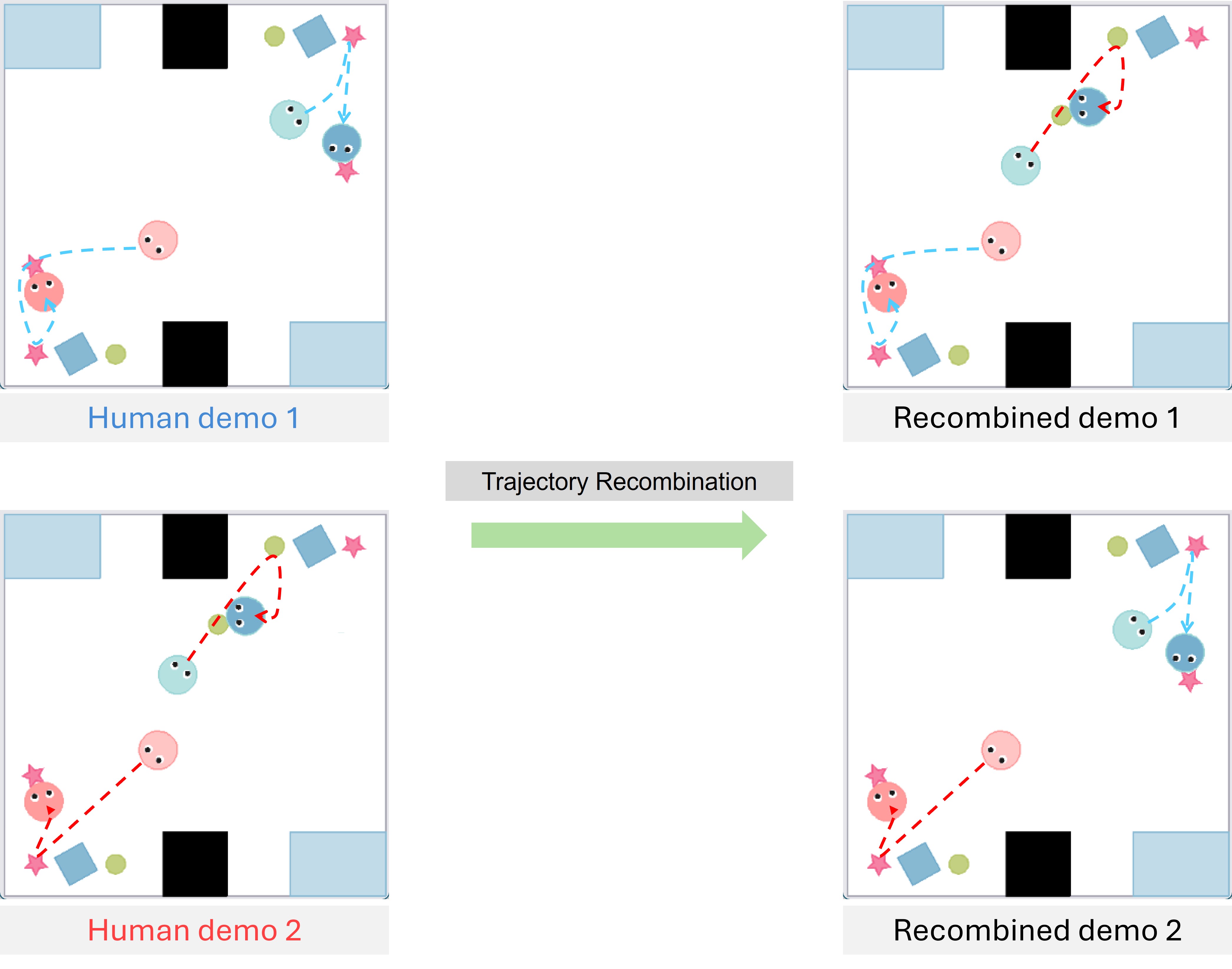}
    \caption{Two human demonstrations (left) contain trajectory segments where the red agent begins from almost identical states, indicating compatible intent and coordination patterns. Based on this compatibility, we exchange the corresponding blue agent segments between the two trajectories to create two new demonstrations (right). This operation preserves trajectory validity while enriching the diversity of collaborative behaviors.}
    \label{fig:recomb_example}
\end{figure*}

Fig. \ref{fig:recomb_example} illustrates an example of our trajectory recombination process. Consider two human demonstrations (or two trajectories in the dataset). If a trajectory segment from the red agent in both demonstrations starts from nearly the same initial state, we treat these segments as behaviorally compatible. This implies that, under this specific behavior of the red agent, the corresponding behaviors of the blue agent in the two demonstrations are mutually acceptable—i.e., they follow a consistent collaboration pattern.

Given this compatibility, the remaining segments from the blue agent in the two trajectories can be exchanged. Replacing the blue agent's segment from demo A with that from demo B (and vice versa) produces two new valid demonstrations. Importantly, these recombined trajectories remain feasible within the environment while significantly increasing the diversity of collaborative behaviors represented in the dataset.

\section{Time-Contrastive Learning as a Reward-Free Progress Estimator}
\label{appx:progress_TCL_esti}
To demonstrate that BASS does not require access to the environment reward, we further evaluate the behavior of the time-contrastive learning (TCL)~\cite{SermanetRewards2017, nair2022rm} model used as a reward-free progress estimator. TCL learns an embedding in which temporal ordering is preserved: earlier frames are embedded closer to the initial frame, while later frames progressively diverge. This structure allows TCL to provide a proxy measure of task progress using only state observations, enabling BASS to operate entirely within an imitation-learning setting.

\begin{figure}[h!]
    \centering
    \includegraphics[width=\textwidth]{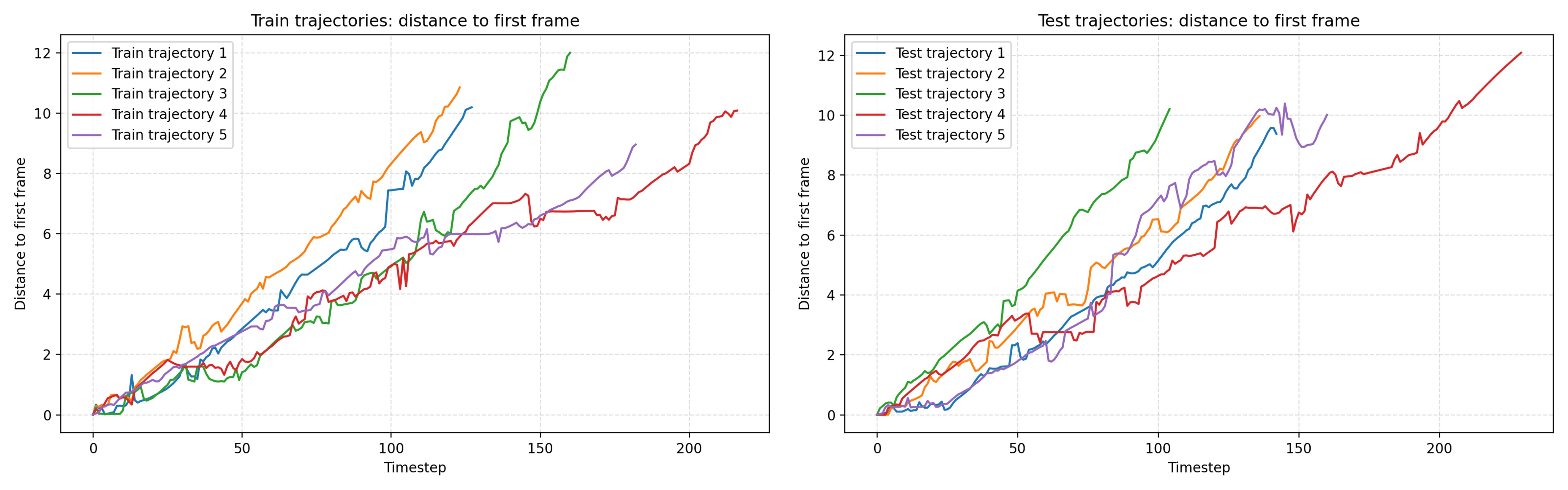}
    \caption{TCL progress estimation. Distances from each frame to the first frame for five training trajectories (left) and five test trajectories (right). The smooth and monotonic increase on both seen and unseen trajectories indicates that TCL provides a consistent and generalizable progress signal without requiring environment rewards.}
    \label{fig:tcl_distance}
\end{figure}

Figure~\ref{fig:tcl_distance} shows the embedding distance between each frame of a trajectory and its first frame. The left panel presents five trajectories from the training set, and the right panel shows five trajectories from the test set. In both cases, the distances increase smoothly as time advances, indicating that TCL captures the underlying notion of task progression. Crucially, the model generalizes to unseen trajectories: the test curves follow patterns similar to the training curves, even though TCL was not trained on these sequences.

\begin{table}[h!]
\centering
\caption{Performance of BASS using a time-contrastive learning (TCL) progress estimator on Moving Out \updatetext{Challenge} 1 and \updatetext{Challenge} 2.}
\label{tab:tcl_results}
\resizebox{0.85\textwidth}{!}{
\begin{tabular}{lcccccccc}
\toprule
& \multicolumn{4}{c}{\textbf{Moving Out \updatetext{Challenge} 1}} & \multicolumn{4}{c}{\textbf{Moving Out \updatetext{Challenge} 2}} \\
\cmidrule(lr){2-5} \cmidrule(lr){6-9}
\textbf{Method} 
& TCR(↑) & NFD(↑) & WT(↓) & AC(↑) 
& TCR(↑) & NFD(↑) & WT(↓) & AC(↑) \\
\midrule
DP/BASS (Ours) 
& 0.3503 & 0.5724 & 0.3598 & 0.8337
& 0.4348 & 0.5535 & 0.3096 & 0.8474 \\

DP/BASS w/ reward estimation 
& 0.3448 & 0.5711 & 0.3475 & 0.8202
& 0.4328 & 0.5411 & 0.3187 & 0.8397 \\

DP 
& 0.3233 & 0.5367 & 0.3789 & 0.8163
& 0.3125 & 0.4526 & 0.3100 & 0.8442 \\
\bottomrule
\end{tabular}
}
\end{table}

This consistency demonstrates that TCL provides a reliable and reward-free progress signal suitable for guiding action selection within BASS. Combined with the quantitative results in Table~\ref{tab:tcl_results}, these findings confirm that the effectiveness of BASS does not rely on having access to environment rewards.

\section{Computing Resources}
\label{appx:computing_resources}

\subsection{Training}

\subsubsection{Behavior Cloning}

Models MLP and GRU are trained for 1000 epochs within approximately 0.5 to 1 hour on a single A6000 GPU. Training a diffusion policy, while also for 1000 epochs, generally requires a longer period of 1 to 3 hours. Overall, the computational time for behavior cloning methods is comparatively short.

\subsubsection{MAPPO}

As MAPPO learns through direct interaction with the environment, it inherently requires a significantly greater number of training iterations. Currently, training MAPPO with 15 CPU threads typically spans 5 to 15 hours. Although MAPPO utilizes a lightweight MLP model with a small number of parameters, its training duration is extended due to two main factors:
\begin{itemize}
    \item Firstly, the simulation environment, which is based on Pymunk, does not support GPU acceleration, thereby limiting the speed of physics calculations and environment stepping.
    \item Secondly, the computation of distance-based rewards becomes a bottleneck, particularly in environments featuring complex wall structures that necessitate more intensive calculations.
\end{itemize}

\subsection{Inference Speed of DP/BASS}
\label{appx:inf_spd_dp_bass}
Inference speed is critical for real-time human-AI collaboration, especially when interacting with human partners. In our setup, the environment runs at 10 Hz, i.e., each step occurs every 100 ms. While diffusion models are generally slower, our implementation generates the next 8 actions in 69 ms on an NVIDIA RTX A6000 GPU. This allows us to interact in real-time by predicting one step in advance – at time step t, the agent executes the action predicted at t–1. This ensures smooth interaction without perceivable lags.

\section{Data Collection: Training Data}
\label{appx:data_coolection}

We conducted data collection for two \updatetext{challenges}, each designed to evaluate different aspects of human-AI collaboration. For the two \updatetext{challenges}, each participant controlled an agent using a joystick. The environment running at 10Hz for data collection.

For \textbf{\updatetext{Challenge} 1}, which focuses on human behavior diversity, we recruited 36 participants, forming 18 groups of two. Before data collection, each group underwent a 10-minute practice session to familiarize itself with the environment. The remaining 50 minutes were dedicated to data collection. Each pair played each map three times, then switched agents and played three more times, resulting in six demonstrations per map. If a group completed all maps, they contributed a total of \(12 \times 6 = 72\) human demonstrations. However, not all groups completed the full set, with some collecting only 3 to 5 demonstrations per map. Additionally, we removed low-quality demonstrations where performance was significantly poor. In total, we collected 1,000 valid human demonstrations for this \updatetext{challenge}.

For \textbf{\updatetext{Challenge} 2}, which evaluates adaptation to physical constraints, we worked with four expert players who were highly familiar with the environment. Each map had randomized object properties, ensuring variation in shape, size, and mass. Each map was played 60 times, resulting in \(60 \times 12 = 720\) human demonstrations.

Our data collection and human study process was approved by an Institutional Review Board (IRB). Participants were compensated based on the amount of data they contributed, receiving between \$15 to \$20 per hour.

\section{Human Study: Playing with Models}
\label{appx:human_study}

\subsection{Human Study Procedure}

To collect data for our project, we designed an interactive experiment where human volunteers collaboratively played with trained AI agents. The data collection process is detailed as follows:

\begin{itemize}
    \item \textbf{Model Selection:} Each volunteer was asked to select a model ID from four provided models (A, B, C, D).

    \item \textbf{Task Description and Limits:} After selecting a model, the volunteer played collaboratively with the AI agent across all twelve maps sequentially. The objective was to move all items on the map into the designated goal region. Each map had a time limit of 50 seconds. The volunteer could proceed to the next map either by successfully moving all items into the goal region or upon reaching the 50-second time limit.

    \item \textbf{Agent Roles:} For the first two models (A and B), the volunteer controlled the "red" agent while the AI controlled the "blue" agent. For the remaining two models (C and D), the roles were switched, with the volunteer controlling the "blue" agent and the AI taking the role of the "red" agent.

    \item \textbf{Questionnaire:} After completing all 12 maps for a given model, the volunteer filled out a questionnaire consisting of eight Likert-scale questions and one free-response question. Responses on the Likert scale ranged from "strongly agree" to "strongly disagree."
\end{itemize}

In total, we conducted this experiment with 32 volunteers. Each volunteer will be paid \$20 for one hour of playing.

\subsection{Questionnaire}
\label{appx:questionnaire}
We use the 7-Point Likert Scale for the questions below:

\begin{enumerate}
    \item \textbf{Teamwork:} The other agent and I worked together towards a goal.
    \item \textbf{Humanlike:} The other agent’s actions were human-like.
    \item \textbf{Reasonable:} The other agent always made reasonable actions throughout the game.
    \item \textbf{Follow:} The other agent followed my lead when making decisions.
    \item \textbf{Physics:} The other agent understands how to work with me when objects have varying physical characteristics.
    \item \textbf{Helpfulness:} The other agent understands my intention and proactively helps me when I need assistance.
    \item \textbf{Collision:} When our movement paths conflict, the other agent and I can effectively coordinate to avoid collisions.
    \item \textbf{Alignment:} When moving large items together, our target directions remain well-aligned.
    \item \textbf{Future:} I would like to collaborate with the other agent in future Moving Out tasks.
\end{enumerate}

\section{Full Results for \updatetext{Challenge} 1 AI-AI Collaboration}
\label{app:task1_full_results}

This section reports the complete experimental results for \updatetext{Challenge} 1 under AI-AI collaboration.

\begin{table}[h]
\centering
\begin{adjustbox}{width=\linewidth}
\begin{tabular}{clcccccccc}
\toprule
\begin{tabular}[c]{@{}c@{}}Evaluation\\ Protocol\end{tabular} & Method  
& TCR (↑) & TCR StdErr 
& NFD (↑) & NFD StdErr 
& WT (↓) & WT StdErr 
& AC (↑) & AC StdErr \\ 
\midrule
\multirow{5}{*}{\begin{tabular}[c]{@{}c@{}}Seen\\ Behaviors\end{tabular}}      
& MLP     & 0.2126 & 0.0072 & 0.2987 & 0.0048 & 0.4896 & 0.0021 & 0.8013 & 0.0093 \\
& GRU     & 0.2369 & 0.0183 & 0.3011 & 0.0142 & 0.4975 & 0.0202 & 0.8151 & 0.0173 \\
& MAPPO   & 0.1929 & 0.0038 & 0.3182 & 0.0045 & 0.5766 & 0.0068 & 0.8097 & 0.0071 \\
& DP      & 0.3233 & 0.0279 & 0.5367 & 0.0151 & 0.3789 & 0.0167 & 0.8163 & 0.0162 \\
& DP/BASS & \textbf{0.3503} & 0.0293 & \textbf{0.5724} & 0.0232 & \textbf{0.3598} & 0.0182 & \textbf{0.8337} & 0.0146 \\ 
\midrule
\multirow{5}{*}{\begin{tabular}[c]{@{}c@{}}Unseen\\ Behaviors\end{tabular}}     
& MLP     & 0.1433 & 0.0061 & 0.2413 & 0.0033 & 0.5647 & 0.0031 & 0.7729 & 0.0090 \\
& GRU     & 0.1638 & 0.0092 & 0.2453 & 0.0026 & 0.5758 & 0.0065 & 0.7830 & 0.0037 \\
& MAPPO   & 0.1635 & 0.0067 & 0.2808 & 0.0037 & 0.6379 & 0.0013 & 0.7858 & 0.0050 \\
& DP      & 0.2563 & 0.0152 & 0.4589 & 0.0177 & 0.4249 & 0.0136 & 0.7854 & 0.0041 \\
& DP/BASS & \textbf{0.3010} & 0.0223 & \textbf{0.5197} & 0.0361 & \textbf{0.3899} & 0.0245 & \textbf{0.8099} & 0.0179 \\ 
\midrule
\multirow{2}{*}{\begin{tabular}[c]{@{}c@{}}Play with\\ Human\end{tabular}} 
& DP      & 0.3855 & 0.0512 & 0.5547 & 0.0432 & 0.4886 & 0.0457 & 0.8054 & 0.0129 \\
& DP/BASS & \textbf{0.6512} & 0.0717 & \textbf{0.7053} & 0.0459 & \textbf{0.3364} & 0.0481 & \textbf{0.9124} & 0.0113 \\ 
\bottomrule
\end{tabular}
\end{adjustbox}
\caption{\updatetext{Challenge} 1 results under seen and unseen human behaviors, and with real human partners.}
\label{tab:task_1_results}
\end{table}

\newpage
\section{Results of Different Methods with BASS}
\label{appx:full_results}
\begin{table}[!h]
\begin{scalebox}{0.9}{
\begin{tabular}{lcccccccc}
\updatetext{Challenge} 1                       & \multicolumn{2}{c}{TCR↑} & \multicolumn{2}{c}{NFD↑} & \multicolumn{2}{c}{WT ↓} & \multicolumn{2}{c}{AC↑}  \\
\multicolumn{1}{c}{Methods} & Mean & Std Error & Mean & Std Error & Mean & Std Error & Mean & Std Error \\ \hline
MLP                          & 0.3568  & 0.0508         & 0.4118  & 0.0338         & 0.4380  & 0.0419         & 0.7890  & 0.0250         \\
+ BASS w/o Simulation                     & 0.2952  & 0.0436         & 0.4207  & 0.0359         & 0.3639  & 0.0358         & 0.8060  & 0.0126         \\ \hline
GRU                          & 0.3070  & 0.0479         & 0.3674  & 0.0365         & 0.3143  & 0.0532         & 0.7618  & 0.0201         \\
+ BASS w/o Simulation                       & 0.4117  & 0.0465         & 0.4396  & 0.0350         & 0.3891  & 0.0418         & 0.8225  & 0.0173         \\
+ BASS w/o Augmentation                   & 0.3531  & 0.0411         & 0.4047  & 0.0373         & 0.3835  & 0.0419         & 0.8195  & 0.0210         \\
+ Full BASS              & 0.4120  & 0.0513         & 0.4454  & 0.0392         & 0.4218  & 0.0426         & 0.8345  & 0.0173         \\ \hline
Diffusion Policy (DP)        & 0.3829  & 0.0681         & 0.4818  & 0.0514         & 0.3075  & 0.0374         & 0.8395  & 0.0216         \\
 + BASS w/o Simulation                      & 0.4028  & 0.0666         & 0.5114  & 0.0493         & 0.3392  & 0.0428         & 0.8242  & 0.0254         \\
 + BASS w/o Augmentation                    & 0.4741  & 0.0667         & 0.5561  & 0.0506         & 0.3176  & 0.0435         & 0.8495  & 0.0174         \\
 + Full BASS               & 0.5027  & 0.0619         & 0.5707  & 0.0468         & 0.3448  & 0.0402         & 0.8615  & 0.0167        
\end{tabular}
\label{tab:full_results_in_t1}
}
\end{scalebox}

\caption{The table presents all experimental results for \updatetext{Challenge} 1 in seen behaviors.}
\end{table}

\begin{table}[!h]
\begin{scalebox}{0.9}{
\begin{tabular}{lcccccccc}
\updatetext{Challenge} 2               & \multicolumn{2}{c}{TCR↑} & \multicolumn{2}{c}{NFD↑} & \multicolumn{2}{c}{WT ↓} & \multicolumn{2}{c}{AC↑}  \\
\multicolumn{1}{c}{Methods}              & Mean & Std Error & Mean & Std Error & Mean & Std Error & Mean & Std Error \\ \hline
MLP                  & 0.2557  & 0.0413         & 0.3602  & 0.0315         & 0.4867  & 0.0418         & 0.8175  & 0.0261         \\
+ BASS w/o Simulation             & 0.2014  & 0.0336         & 0.3656  & 0.0244         & 0.3657  & 0.0332         & 0.7890  & 0.0250         \\ \hline
GRU                  & 0.2582  & 0.0509         & 0.3935  & 0.0428         & 0.4680  & 0.0594         & 0.8487  & 0.0183         \\
+ BASS w/o Simulation               & 0.3333  & 0.0539         & 0.4141  & 0.0439         & 0.5611  & 0.0587         & 0.8513  & 0.0286         \\
+ BASS w/o Augmentation           & 0.3670  & 0.0522         & 0.4246  & 0.0420         & 0.4365  & 0.0593         & 0.8572  & 0.0222         \\
+ Full BASS      & 0.3414  & 0.0522         & 0.4410  & 0.0442         & 0.4379  & 0.0596         & 0.8754  & 0.0165         \\ \hline
Diffusion Policy (DP) & 0.3125  & 0.0564         & 0.4526  & 0.0427         & 0.3100  & 0.0385         & 0.8442  & 0.0184         \\
+ BASS w/o Simulation              & 0.3569  & 0.0547         & 0.4908  & 0.0385         & 0.3256  & 0.0431         & 0.8373  & 0.0147         \\
+ BASS w/o Augmentation            & 0.4200  & 0.0544         & 0.5187  & 0.0417         & 0.3232  & 0.0417         & 0.8305  & 0.0169         \\
+ Full BASS       & 0.4348  & 0.0599         & 0.5535  & 0.0423         & 0.3096  & 0.0451         & 0.8474  & 0.0128        
\end{tabular}
}
\end{scalebox}
\label{tab:full_results_in_t2}
\caption{The table presents all experimental results for \updatetext{Challenge} 2}
\end{table}

% \newpage
\section{Map Analysis}
\label{appx:all_maps}

Our 12 maps are carefully designed to target specific collaboration modes, coordination, awareness, and action consistency, while ensuring that existing AI agents (e.g., MAPPO, Diffusion Policy) can perform some tasks but still exhibit clear limitations. This balance is essential: overly difficult maps with long paths or dense obstacles may yield near-zero performance for all agents, making it impossible to evaluate various aspects of human-AI collaboration. Our map definition is already based on a structural format, e.g., JSON, allowing easy modification, reuse of modules, and procedural generation for scalability.

\renewcommand{\arraystretch}{1.9}

\subsection{Coordination}

    \begin{table}[!h]
    \centering
    \begin{tabular}{|m{3cm}|m{3.5cm}|m{3cm}|m{3.5cm}|}
    \hline
    \textbf{Map} & \textbf{Analysis} & \textbf{Map} & \textbf{Analysis} \\ \hline
    
    \includegraphics[width=3cm]{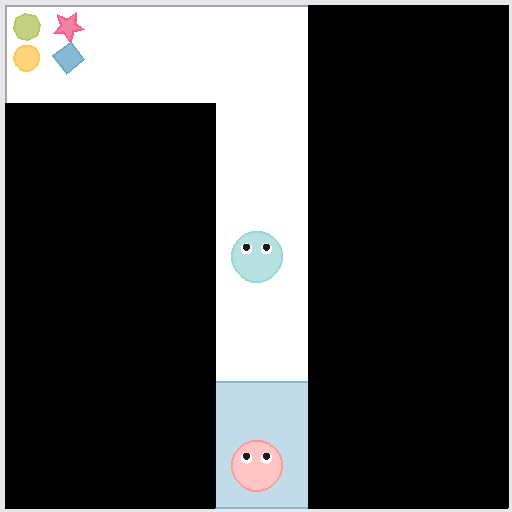} & \textit{Map 1: Hand Off} is designed with a single narrow pathway that forces the agent, the one that is closer to the items, to efficiently pass them to the other agent.  
    & \includegraphics[width=3cm]{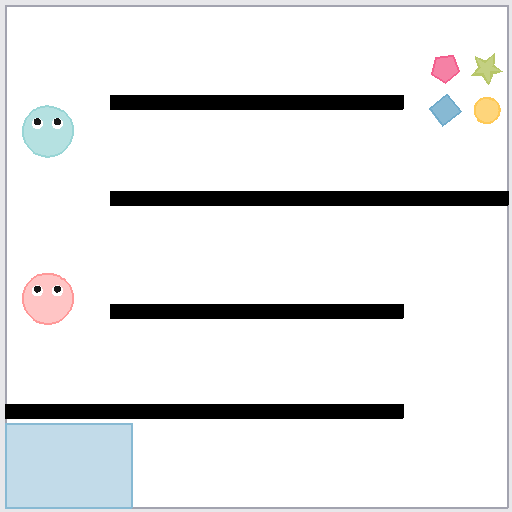} & \textit{Map 2: Pass Or Split} features four non-intersecting pathways, designed to evaluate the agents' ability to select the most suitable path while considering the need for collaboration. \\ \hline
    
    \includegraphics[width=3cm]{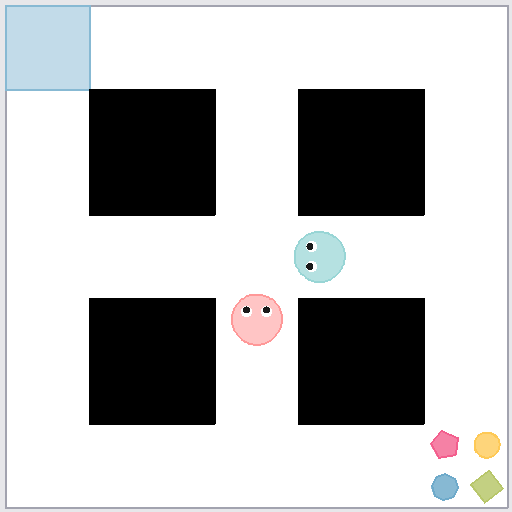} & \textit{Map 3: Efficient Routes} features several pathways leading to the goal region, allowing the agents to independently determine the most efficient path while considering the movement of the other agent.  
    &  \includegraphics[width=3cm]{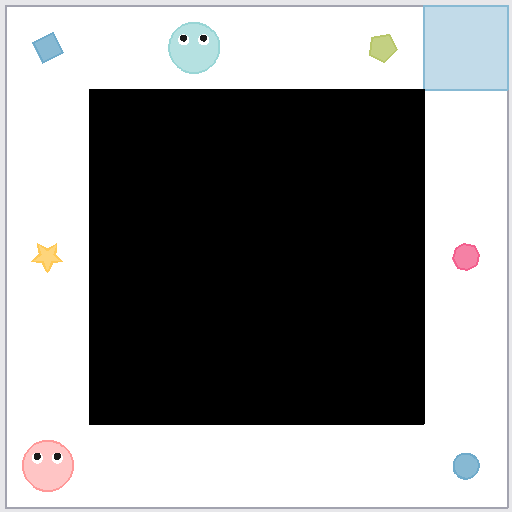} & \textit{Map 4: Priority Pick} creates an environment that requires each agent to independently decide whether to prioritize moving the item closer to the goal region first or bringing the farther item closer. \\ \hline
    \end{tabular}
    % \vspace{-1em}
    \caption{Maps categorized under \textbf{Coordination}.}
    \label{tab:maps_coordination}
    \end{table}

% \newpage

\subsection{Awareness}

    \begin{table}[!h]
    \centering
    \begin{tabular}{|m{3cm}|m{3.5cm}|m{3cm}|m{3.5cm}|}
    \hline
    \textbf{Map} & \textbf{Analysis} & \textbf{Map} & \textbf{Analysis} \\ \hline
    \includegraphics[width=3cm]{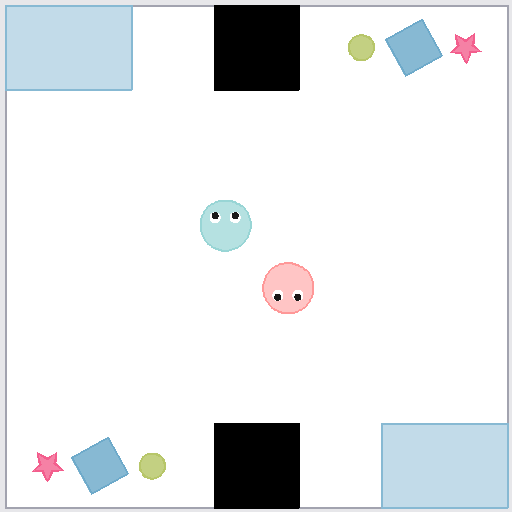} & \textit{Map 5: Corner Decision} requires the agents to decide whether to follow the other agent to the upper right or the lower left corner and to determine which size of item to prioritize moving first.    
    & \includegraphics[width=3cm]{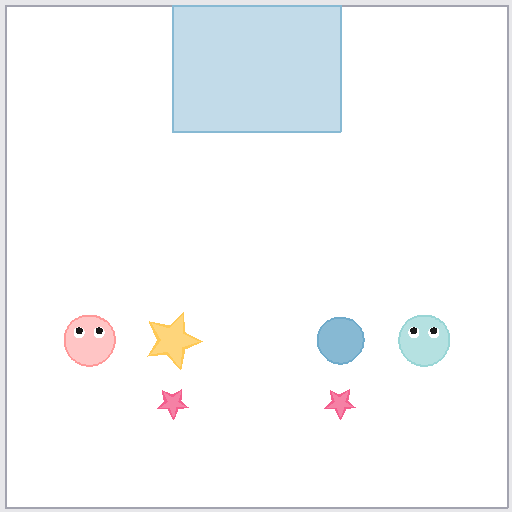} & \textit{Map 6: Distance Priority} contains two medium-sized items, requiring the agents to decide whether to prioritize the item that is farther away or the one that is closer. \\ \hline
    
    \includegraphics[width=3cm]{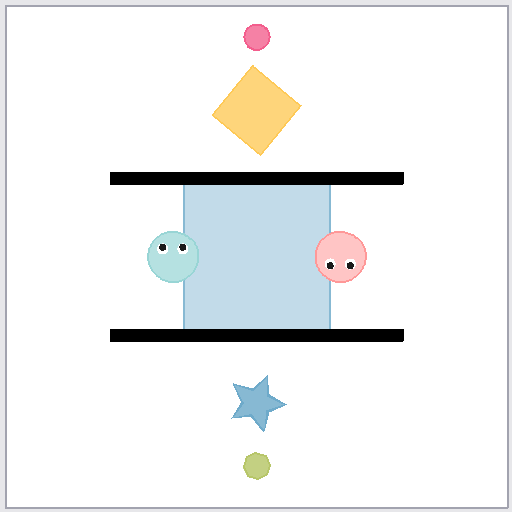} & \textit{Map 7: Top Bottom Priority} contains two items, either large or medium-sized, requiring the agents to decide whether to prioritize the item at the top or the one at the bottom. 
    & \includegraphics[width=3cm]{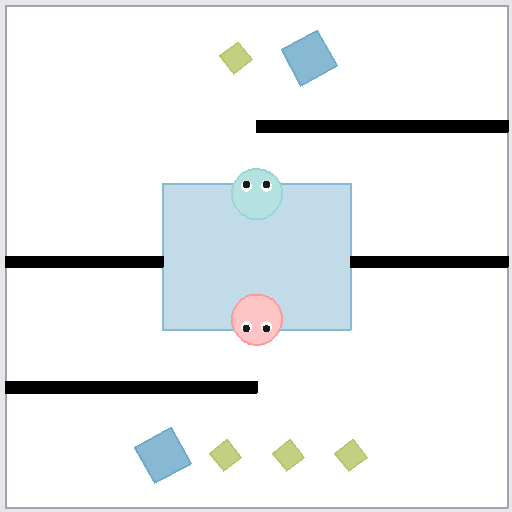} & \textit{Map 8: Adaptive Assist} contains a mix of large or medium-sized items and small items, requiring the agents to decide whether to prioritize collaborating on the larger item or individually handling the smaller item. \\ \hline
    \end{tabular}
    % \vspace{-1em}
    \caption{Maps categorized under \textbf{Awareness}.}
    \label{tab:maps_awareness}
    \end{table}
    \newpage
    % \label{appx:all_maps}
    \renewcommand{\arraystretch}{1.9}

\subsection{Action Consistency}

    \begin{table}[!h]
    \centering
    \begin{tabular}{|m{3cm}|m{3.5cm}|m{3cm}|m{3.5cm}|}
    \hline
    \textbf{Map} & \textbf{Analysis} & \textbf{Map} & \textbf{Analysis} \\ \hline
    \includegraphics[width=3cm]{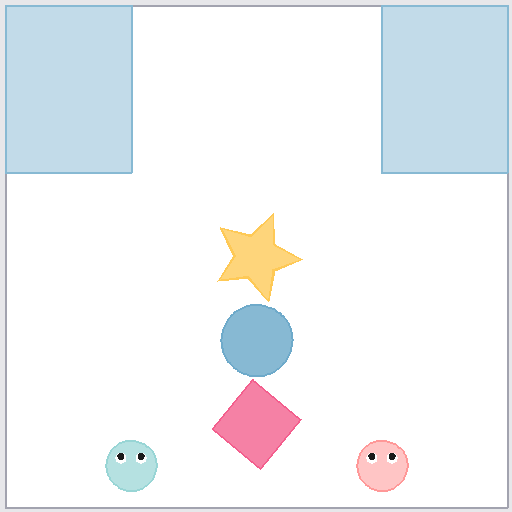} & \textit{Map 9: Left Right} contains large-sized items, requiring the agents to continuously collaborate and make strategic decisions about whether to move items to the left or right goal region.  
    & \includegraphics[width=3cm]{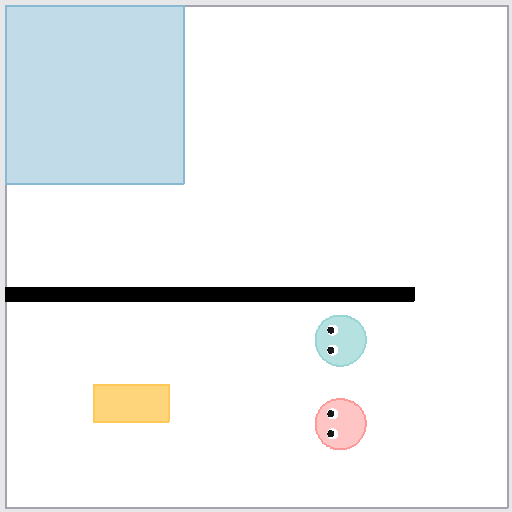} & \textit{Map 10: Single Rotation} contains one large-sized item, which is designed to evaluate how well the two agents can collaborate to perform a single rotation.    \\ \hline

    \includegraphics[width=3cm]{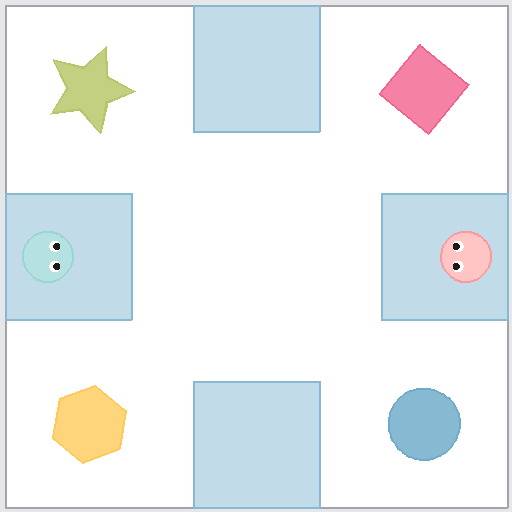} & \textit{Map 11: Four Corners} contains large-sized items positioned at the four corners, requiring the agents to continuously collaborate by moving the items in either a clockwise or counter-clockwise order.
    & \includegraphics[width=3cm]{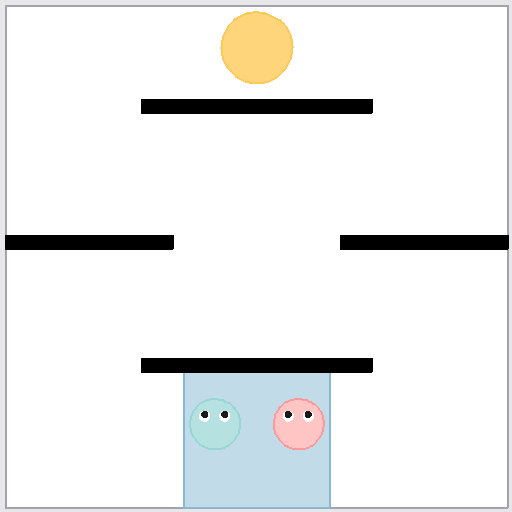} & \textit{Map 12: Sequential Rotations} contains one large-sized item, which is designed to evaluate how well the two agents can collaborate to maintain a sequence of rotations. \\ \hline
    \end{tabular}
    % \vspace{-1em}
    \caption{Maps categorized under \textbf{Action Consistency}.}
    \label{tab:maps_action_consistency}
    \end{table}

\end{document}